\documentclass[lettersize,journal]{IEEEtran}
\usepackage{amsmath,amsfonts}
\usepackage{algorithmic}
\usepackage{algorithm}
\usepackage{array}
\usepackage[caption=false,font=normalsize,labelfont=sf,textfont=sf]{subfig}
\usepackage{textcomp}
\usepackage{stfloats}
\usepackage{url}
\usepackage{verbatim}
\usepackage{graphicx}
\usepackage{cite}

\begin{document}

\title{TADIL: Task-Agnostic Domain-Incremental Learning through Task-ID Inference using Transformer Nearest-Centroid Embeddings}

\author{Gusseppe Bravo-Rocca, Peini Liu, Jordi Guitart, Ajay Dholakia, David Ellison
\thanks{G. Bravo-Rocca, P. Liu, and J. Guitart are with the Barcelona Supercomputing Center, Barcelona, Spain (e-mail: gusseppe.bravo@bsc.es; peini.liu@bsc.es; jordi.guitart@bsc.es).}
\thanks{J. Guitart is also with the Universitat Polit\`ecnica de Catalunya, Barcelona, Spain.}
\thanks{A. Dholakia and D. Ellison are with Lenovo Infrastructure Solutions Group, Morrisville, NC, USA (e-mail: adholakia@lenovo.com; dellison@lenovo.com).}
}

% \author{IEEE Publication Technology,~\IEEEmembership{Staff,~IEEE,}
%         % <-this % stops a space
% \thanks{This paper was produced by the IEEE Publication Technology Group. They are in Piscataway, NJ.}% <-this % stops a space
% \thanks{Manuscript received April 19, 2021; revised August 16, 2021.}}

% The paper headers
 \markboth{IEEE TRANSACTIONS ON PATTERN ANALYSIS AND MACHINE INTELLIGENCE, VOL. X, NO. X, JUN 2023}%
{G. Bravo-Rocca \MakeLowercase{\textit{et al.}}: TADIL: Task-Agnostic Domain-Incremental Learning through Task-ID Inference using Transformer Nearest-Centroid Embeddings}

% \markboth{Journal of \LaTeX\ Class Files,~Vol.~14, No.~8, August~2021}%
% {Shell \MakeLowercase{\textit{et al.}}: A Sample Article Using IEEEtran.cls for IEEE Journals}

% \IEEEpubid{0000--0000/00\$00.00~\copyright~2021 IEEE}
% Remember, if you use this you must call \IEEEpubidadjcol in the second
% column for its text to clear the IEEEpubid mark.

\maketitle

\begin{abstract}
Classical Machine Learning (ML) models struggle with data that changes over time or across domains due to factors such as noise, occlusion, illumination, or frequency, unlike humans who can learn from such non independent and identically distributed data. Consequently, a Continual Learning (CL) approach is indispensable, particularly, Domain-Incremental Learning. %, as classical static ML approaches are inadequate to deal with data that comes from different distributions.
In this paper, we propose a novel pipeline for identifying tasks in domain-incremental learning scenarios without supervision. The pipeline comprises four steps. First, we obtain base embeddings from the raw data using an existing transformer-based model. Second, we group the embedding densities based on their similarity to obtain the nearest points to each cluster centroid. Third, we train an incremental task classifier using only these few points. Finally, we leverage the lightweight computational requirements of the pipeline to devise an algorithm that decides in an online fashion when to learn a new task using the task classifier and a drift detector.
We conduct experiments using the SODA10M real-world driving dataset and several CL strategies. We demonstrate that the performance of these CL strategies with our pipeline can match the ground-truth approach, both in classical experiments assuming task boundaries, and also in more realistic task-agnostic scenarios that require detecting new tasks on-the-fly. 
\end{abstract}


\section{Introduction}
\label{sec:intro}
\IEEEPARstart{T}{he} field of Machine Learning (ML) has made significant strides in recent years, thanks to the availability of vast amounts of data. However, classical approaches assume that the data used for training and inference is independent and identically distributed (IID), which is not always true in real-life applications. In reality, data can be non-IID, correlated, and present in different contexts, which leads to the domain shift problem, which means that the data distribution changes across tasks or classes. Furthermore, when a model is deployed for inference, it typically assumes that there will be no distribution drifts over time, which makes it convenient in implementation but restrictive for real applications.

% JG: Is the drift checked for each individual input x, or for a batch of inputs?
% gb: for a batch of inputs
% JG: then, in the caption of the figure, the input 'x' refers to a batch of inputs?
% GB: Yes. The argument for that is because in real world we might trigger the approach for every bunch of frames instead of one by one which would be costly.
% JG: ok, but then we should probably clarify that when explaining the figure
% GB: DONE
% JG: ok, but when you refer to task $T$ and $\hat T$, it should be $T_i$ and $\hat T_i$, right?
% GB: Yep
% JG: Figure must be refined to emphasize our approach, and to match better with the description in later sections
% GB: Done
\begin{figure}[!t]
  \centering
   \includegraphics[width=0.9\linewidth]{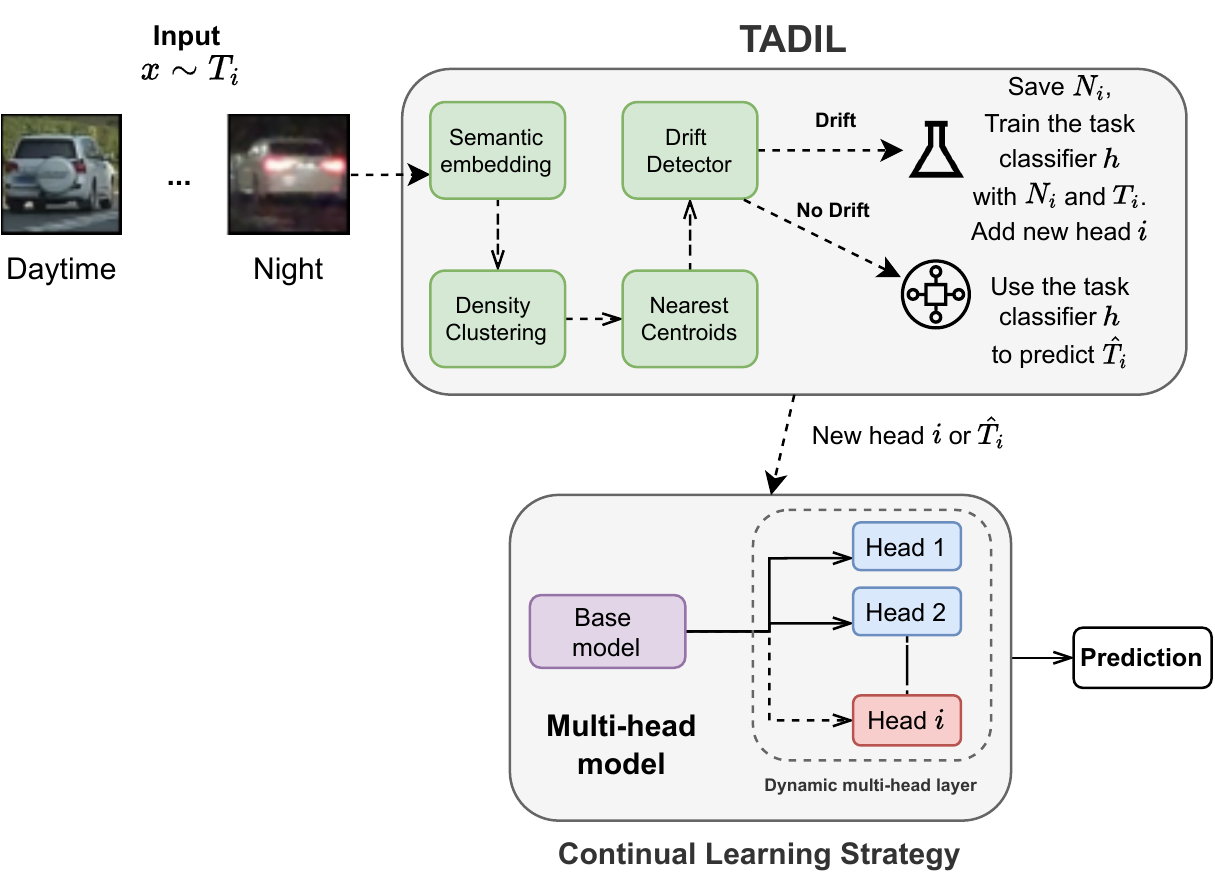}

   \caption{Given a batch of images as input $x$ associated with a specific task $T_t$, our method calculates the nearest-centroid embeddings $N_t$ and then checks whether they exhibit drift. If drift is present, we save $N_t$ in memory $\mathcal{M}$ and then a task classifier $h$ is incrementally trained using $N_t$ and $T_t$ with no supervision. Conversely, if no drift is detected, the classifier $h$ is employed to estimate the task $\hat T_t$. The multi-head classifier then selects the appropriate classifier based on the predicted task, generating the final prediction $y$.}
   \label{fig:intro}
\end{figure}

To address these limitations, the Continual Learning (CL) field seeks to develop algorithms that do not stop learning even after deployment, effectively unfreezing the training phase and adapting it continuously while the model is used for inference. However, saving all the incoming data and using it for CL is not feasible due to privacy, storage, and computational resource constraints.

Several methods have been developed to address those issues, namely regularization \cite{Zenke2017}, which aims to prevent the model parameters from changing too much when learning new tasks; replay \cite{Lopez-Paz2017}, which involves storing and revisiting some data from previous tasks to avoid forgetting; and architecture modifications \cite{Rebuffi2017,Aljundi2018}, which involve adding or pruning neurons or layers to adapt to new tasks. However, these approaches typically assume rigid task boundaries and known tasks, which means that they require explicit signals or labels to switch between tasks or classes. This assumption may not hold in many real-world scenarios where the data is continuous and heterogeneous with no supervision.

To overcome these limitations, we devise a novel task-agnostic approach for domain-incremental learning that can detect task drift and classify incoming tasks in an unsupervised manner. The drift detection is able to recognize any new domain in a data stream without requiring any labels, by monitoring the changes in the data distribution over time. The classification is able to learn to assign a unique task ID to each domain, by exploiting the similarities and differences among the data samples from different domains. The task ID can then be used by the CL strategies that rely on multi-head models, which have a separate output head for each task. These models can be easily created by adding a multi-head layer on top of the base model, which allows the core of the model to remain unchanged. In this way, we can segment the data stream into meaningful domains and classes without any supervision, and apply appropriate CL strategies to each domain to adapt and optimize their parameters according to the specific task at hand, further improving their performance. To the best of our knowledge, no other works have proposed unsupervised approaches for identifying and classifying tasks in task-agnostic domain-incremental learning scenarios for driving datasets.

% GB: The following paragraph may need some touch-ups to explain better the approach
Our approach is implemented as a pipeline for detecting and identifying tasks in domain-incremental learning scenarios without supervision, as shown in Fig.~\ref{fig:intro}. The pipeline consists of four primary steps. For each batch of inputs, it leverages an existing transformer-based model to obtain a base embedding from raw data. Then, it groups the embedding densities based on their similarity to obtain the nearest points to each cluster centroid and trains an incremental task classifier using only these points. Finally, thanks to the lightweight computational requirements of the pipeline, we use it to devise an algorithm that can decide in an online fashion when to learn a new task using the task classifier and a drift detector. We evaluate our approach using the SODA10M real-world driving dataset on a CPU-based platform (no GPU is required due to the low computational requirements of our solution) and demonstrate that it is beneficial when performing inference with state-of-the-art CL strategies based on multi-head classifiers.

\subsection{Contributions}
\noindent The main contributions of this paper can be summarized as follows:

\begin{itemize}
    \item We propose an unsupervised approach for identifying and classifying tasks in domain-incremental learning scenarios based on transformed nearest-centroid semantic embeddings.
    
    \item We implement our domain-incremental learning approach as a pipeline that uses semantic embeddings, density-based clustering, and nearest-cluster centroids to train incrementally a drift detector and a task classifier that can predict if a new task appears in inference time and the task label for a given input, respectively.

    \item We devise an online algorithm that uses the proposed drift detector and task classifier to decide when to learn a new task and feed a multi-head model with the adequate task ID in task-agnostic scenarios.

    \item We evaluate our domain-incremental learning approach on several state-of-the-art CL strategies by conducting experiments using the SODA10M real-world driving dataset in scenarios with task boundaries and task-agnostic.
\end{itemize}

The remainder of the paper is as follows: Section \ref{sec:related} starts with a review of related work, focusing on the problem of catastrophic forgetting in CL, and then, it introduces two specific types of CL problems, namely Domain-incremental learning and Task-Agnostic Continual Learning (TACL). Section \ref{sec:definition} presents the problem definition in terms of input space, output space, and task IDs. Section \ref{sec:approach} outlines our approach to train a task classifier in an unsupervised way, which involves several components such as semantic embeddings, density-based clustering, nearest-cluster centroids, nearest-centroid incremental classifier, and drift detector. Section \ref{sec:pipeline} describes an online pipeline algorithm to infer the task IDs using the previous approach. Section \ref{sec:experiments} presents an experimental evaluation in both task-boundary and task-agnostic setups. Finally, Section \ref{sec:conclusions} concludes the paper with a discussion of the results and potential future work.

\section{Related work}
\label{sec:related}

\noindent This section introduces some related works in the field of Continual Learning which are relevant to the research presented in this paper.

\subsection{Multimodal transformers}
% GB: explain here the use of multimodal transformer for CL
% GB: Done
\noindent Multimodal transformers, such as the CLIP \cite{radford2021learning} model, have emerged as an important component in CL. These models are designed to process and learn from both visual and textual inputs, which allow them to generate rich embeddings for images, capturing essential features and semantic information. These embeddings enable the CL models to distinguish better between diverse objects and situations, while also supporting the CL process by making it more resilient to novel situations. This results in a more robust and efficient learning process when applied to the complex task of autonomous driving. 
% GB: add related work btw CLIP and CL
% GB: Done

For instance, TransFuser \cite{fusion_transformer} is a multi-modal Fusion Transformer that integrates image and LiDAR representations using attention. It has been experimentally validated in urban settings involving complex scenarios using the CARLA urban driving simulator. Furthermore, Huang et al. \cite{motion_transformer} introduced a neural prediction framework utilizing the Transformer structure, with a multi-modal attention mechanism for representing social interactions between agents and predicting multiple trajectories for autonomous driving. In our approach, we do not train a transformer-based model from scratch, as this would be too expensive to do for each application. We think that existing large transformer-based models are sufficient to obtain general patterns of our environment. Thanks to their modularity, these models may be replaced by better versions if needed.

%-------------------------------------------------------------------------
\subsection{Catastrophic forgetting}
\noindent One of the biggest problems in CL is catastrophic forgetting, which is an issue that makes neural networks forget what they have learned when acquiring new concepts \cite{Delange_2021}, mostly due to gradient descent \cite{Kirkpatrick_2017}. In order to deal with both past and new data, we should weigh between integrating novel data (plasticity) and not interfering with learned knowledge (stability), the so-called \textit{stability-plasticity} dilemma. This follows the biological mechanism that occurs in humans; the hippocampal system exhibits short-term adaptation (rapid learning, that is, plasticity) whereas the neocortical system holds the long-term storage (slow learning, generalities, that is, stability) \cite{PARISI201954}. 
Several approaches have been proposed to tackle this problem, although it is still an open challenge. For instance, Kirkpatrick et al. \cite{Kirkpatrick_2017} presented the Elastic Weight Consolidation (EWC) strategy that performs a protection of old knowledge during new learning by decreasing the plasticity of weights (by applying regularization). This is inspired in how the mammalians brains perform a synaptic consolidation (past synapses are strengthened) to persist old skills when learning other tasks. Rolnick et al. \cite{rolnick19} described the Experience Replay (ER) strategy, a rehearsal-based method that addresses forgetting by maintaining a memory buffer of previously learned experiences to then combine with new data.
%JG: Here, we must add references to Experience Replay, and Learning without Forgetting (as we did with EWC)
% JG: done

%-------------------------------------------------------------------------

\subsection{Domain-incremental learning}

\noindent Domain-incremental learning (Domain-IL) is a kind of CL problem that focuses on learning multiple tasks sequentially where each task has its own domain. In the context of driving datasets, domain-incremental learning can be applied to improve continuously the performance of models that predict driving behavior, such as detecting pedestrians, vehicles, road signs, etc., even when the car changes to a different domain (i.e., different location, time of the day, or weather conditions). 

Several methods have been proposed for domain-incremental learning in machine learning, including DISC (Domain Incremental through Statistical Correction), which is an online zero-forgetting approach that can incrementally learn new tasks without requiring re-training or expensive memory banks \cite{meng2022discl}. DISC expects the task ID at inference time, which is obtained by using physical sensors. Conversely, in our approach, we learn the task ID. Similarly, González et al.~\cite{cl_medical_seg} train an autoencoder network for each task to identify the domain during inference. When an image arrives at inference time, they calculate the reconstruction error against each of the autoencoders. The task ID of the autoencoder with the smallest error is chosen. However, training one autoencoder for each domain is time-consuming. Besides, a task boundary is expected, which might not be feasible for other scenarios. Our method only requires training a lightweight classifier with a few examples in a task-agnostic way (no boundaries).
Finally, domain-aware categorical representations are another method for general incremental learning \cite{xie2022general}. An offline framework with a flexible class representation based on a mixture model is used to address the stability-plasticity dilemma and imbalance challenges. It is interesting to see how drift is handled in imbalance problems, however, this requires an internal modification of the model. In our work, we seek to infer the task ID from outside of the model.

%-------------------------------------------------------------------------
\subsection{Task-Agnostic Continual Learning (TACL)}

\noindent TACL is a type of lifelong learning that aims to continuously learn from non-stationary distributions, where the identity of tasks is unknown at training time \cite{shin2022task}. CL with driving datasets is particularly relevant here as driving scenarios are constantly evolving and require models to continuously adapt to new conditions. Several studies have explored task-agnostic CL in the context of driving datasets, using various techniques to mitigate catastrophic forgetting and improve model performance. Shin et al. \cite{Shin_2017} used generative replay to generate data from previous tasks during the training of new tasks to prevent forgetting, thereby improving performance on both past and new tasks. However, generative models are costly in terms of resources, therefore, limited to offline applications. The Learning without Forgetting (LwF) strategy proposed by Li et al. \cite{li2018learning} is an architecture-based method that prevents forgetting while retaining the performance on previous tasks by using a combination of distillation and gradient-based regularization so that the model's own predictions on the new task data are used as 'soft targets' for the old tasks during training. In contrast, Schwarz et al. \cite{schwarz2018progress} proposed a method called 'progress and compress' that combines weight consolidation with lifelong generative models to achieve task-agnostic CL. The task IDs are not given either in the work by Rebuffi et al. \cite{Rebuffi2017}, which targets class-incremental learning. Only the training data for a small number of classes is present at the same time, instead of having all class data available at once. 

The aforementioned approaches for domain-incremental learning and task-agnostic continual learning, while effective in their respective contexts, suffer from some significant drawbacks. One primary limitation is that they are not model agnostic, meaning that the techniques are specifically designed for particular models and may not generalize well across different model architectures. This lack of flexibility could hinder their applicability in real-world scenarios, where diverse models are often needed to address various problems effectively. Additionally, these approaches rely on the same model to both address the main problem, such as classification or object detection, and to detect task boundaries. This dual-purpose design can introduce bias in the learning process, as the model may be influenced by its own structure and assumptions when attempting to identify and adapt to new tasks. Consequently, the model's ability to effectively learn and generalize to new tasks may be compromised, resulting in sub-optimal performance. In this paper, we use a more robust approach, which leverages a lightweight, independent model that can be easily integrated into various architectures, allowing for unbiased task identification and adaptation while retaining performance on previous tasks.

\subsection{Other challenges for CL}
\noindent CL is a complex problem that involves not only mitigating forgetting but also addressing other challenges such as learning efficiency, hardware resource utilization, backward and forward transfer, and robustness. To deal with these issues, a number of strategies have been suggested. Notably, as with catastrophic forgetting, those are still ongoing challenges and further research in these topics is required. For instance, Aljundi et al. \cite{Aljundi_2019} proposed an online meta-learning approach that learns a prior distribution over model weights, which can be used to quickly adapt to new tasks. 
% JG: \cite{Shin_2017} is also cited in 2.2, so I comment here
%A technique for selective replay was described by Shin et al. \cite{Shin_2017}, which leverages a generative model to generate data for past tasks during training on new tasks, thereby improving performance on both past and new tasks. 
Other approaches are based on dynamic architectures \cite{rusu2016progressive}, where the structure of the model evolves during training (e.g., adding more layers dynamically), allowing it to adapt and improve its performance; distillation \cite{li2017learning}, which involves training a smaller, more efficient model to mimic the behavior of a larger, pre-trained one, thus, retaining its knowledge while reducing complexity; and attention-based models \cite{mallya2018packnet}, which use attention mechanisms to focus on relevant information, enhancing the ability of the model to process and learn from complex data.

%-------------------------------------------------------------------------
\section{Problem definition}
\label{sec:definition}
\noindent Let $X$ be the input space, $Y$ be the output space, and $T$ be the space of task IDs. We consider a sequence of $K$ tasks, where each task $k$ is associated with a joint distribution $P_k(X,Y)$ over $X \times Y$. The goal of our domain-incremental learning approach is to learn a sequence of $K$ models $f_1, f_2, ..., f_K$ (in our case, these will be classifier models), where $f_k: X \to Y$ is the model for task $k$ (they are not necessarily different from each other), such that each model can be learned incrementally from data without forgetting the previous tasks, i.e., when learning $f_k$, the models $f_1, f_2, ..., f_{k-1}$ should be preserved.

During inference, the task ID $t \in T$ will be unknown. The model $f_t$ for each task $t$ will be used to predict the output $y \in Y$, that is, $f_t = p_{t}(y|x)$. More formally, we can define our domain-incremental learning approach as:

\begin{equation}
\arg\min_{f_1, f_2, ..., f_K} \sum_{k=1}^K \mathcal{L}(f_k, P_k)
\end{equation}
where the goal is to minimize the loss function $\mathcal{L}(f_k, P_k)$ over a set of $K$ functions $f_1, f_2, ..., f_K$, subject to the constraint that each of the $K$ functions can be learned incrementally. In other words, the functions can be updated or trained on new data without forgetting the knowledge they previously learned.

As for our experiments, we consider the set of functions ${f_1, f_2, \dots, f_K}$ as multi-head classifiers in the following way:

\begin{equation}
f_k(\mathbf{x}) = g_k(e(\mathbf{x})), \quad k=1,2,\dots,K
\end{equation}

where $e(\mathbf{x})$ denotes the shared feature extractor network, $g_k(\cdot)$ denotes the classifier for the $k$-th task, and $\mathbf{x}$ denotes the input sample. Particularly, we use the ResNet18 \cite{he2016deep} architecture as the feature extractor network and linear classifiers on top of it.

% JG: EWC, GEM, AGEM, UCL, Replay appear for the first time here. At some point we need to explain them
% GB: Yes, in progress...
% JG: see my previous comment about adding some refs to the related work
% JG: done
To improve the performance of the multi-head classifier at inference time and enable strategies that require the task ID, such as EWC, ER, LwF, among others, it is necessary to have a task classifier that can learn the task ID with no supervision. This task classifier should take the input sample $\mathbf{x}$ and predict the corresponding task ID $t \in T$, which can then be used to select the appropriate classifier $f_t$ for inference. Without this task classifier, the multi-head classifier may not be able to effectively utilize the knowledge learned from previous tasks and may suffer from catastrophic forgetting. Additionally, having a task classifier that can automatically learn the task ID without supervision can simplify the overall learning process and reduce the amount of manual intervention required.

Let $g_t$ be the classifier for task $t$ learned from data. We define a task classifier $h_t: X \to T$ that takes an input sample $\mathbf{x} \in X$ and predicts the corresponding task ID $t \in T$. This task classifier can be learned without supervision, as it simply needs to predict the correct task ID associated with each input sample.

During inference, given an input sample $\mathbf{x} \in X$ and the predicted task ID $\hat{t} = h(\mathbf{x})$, the multi-head classifier $f_{\hat{t}}$ is used to predict the output $y \in Y$. Thus, the final prediction can be written as:

\begin{equation}
y = f_{\hat{t}}(\mathbf{x}) = g_{\hat{t}}(e(\mathbf{x}))
\end{equation}

Therefore, incorporating a task classifier into the domain-incremental learning framework can be expressed as:

\begin{equation}
\arg\min_{h, g_1, g_2, ..., g_K} \sum_{k=1}^K \mathcal{L}(f_k, P_k)
\end{equation}

where $h$ is the task classifier, $g_k$ is the classifier for task $k$, and $\mathcal{L}(f_k, P_k)$ is the loss function for task $k$. This objective function ensures that each of the $K$ classifiers can be learned incrementally without forgetting the knowledge they previously learned, while also incorporating the task classifier into the learning process.

%------------------------------------------------------------------------
%P: 'Training Pipeline and Drift Detector for Task-agnostic Domain-incremental Learning'? Here we first use a training pipeline to train a classifier and also then train a drift detector, right?
\section{Components of the pipeline for task-agnostic domain-incremental learning}
\label{sec:approach}
%P: Maybe it is better to clearly show these two contributions? At first glance, I thought the detector belonged to the training pipeline. 
%JG: well, both the classifier and the detector are trained online and incrementally, so I don't know if we can distinguish the training and the inference pipelines
%P: In this section, we first present a training pipeline for task-agnostic domain-incremental learning. A series of components are introduced to obtain the nearest centroids and to train an incremental classifier using the Nearest Centroid Algorithm. Then, a drift detector is presented to detect when to train incrementally the classifier.
%JG: ok to adding an introductory paragraph

\noindent In this section, we first present a training pipeline for task-agnostic domain-incremental learning. A series of components are introduced to obtain the nearest centroids and to train an incremental task classifier using the Nearest Centroid Algorithm. Then, a drift detector is presented to detect when to train incrementally the task classifier.

To train the task classifier in an unsupervised way we need a series of components that together can predict the task at inference time. Let $h_t$ be the task classifier for each task $T_t$. The task classifier is a function that maps an input $x \in \mathcal{X}$ to a task label $t \in {1,2,...,T}$, i.e., $h_t: \mathcal{X} \rightarrow {1,2,...,T}$. The task classifier is obtained through the following pipeline:

%JG: when describing the experiments, remember to mention the $f_{emb}$ function used
% GB: Ok
% JG: Now, I think we better mention the $f_{emb}$ function used here
% JG: I added CLIP ViT-B/32 and a reference
\textbf{Semantic embedding}. Given a batch of inputs $X={x_1,x_2,...,x_m}$, where $m$ is the batch size, we first obtain their corresponding embeddings $E={e_1,e_2,...,e_m}$ using the pretrained transformer-based model CLIP ViT-B/32 \cite{radford2021learning}. We can represent this process as $E=f_{emb}(X)$, where $f_{emb}$ is the embedding function. 
The use of a pretrained transformer-based model can be justified by the fact that these models have already been trained on large amounts of data, and as a result, have learned representations of common things such as pedestrians, cars, buildings, and other objects that are commonly found in driving scenarios. Moreover, they capture higher-level semantic information about the input that can be useful for various downstream tasks, such as classification, clustering, or retrieval.

\begin{figure}[t]
  \centering
  % \fbox{\rule{0pt}{2in} \rule{0.9\linewidth}{0pt}}
   \includegraphics[width=1.0\linewidth]{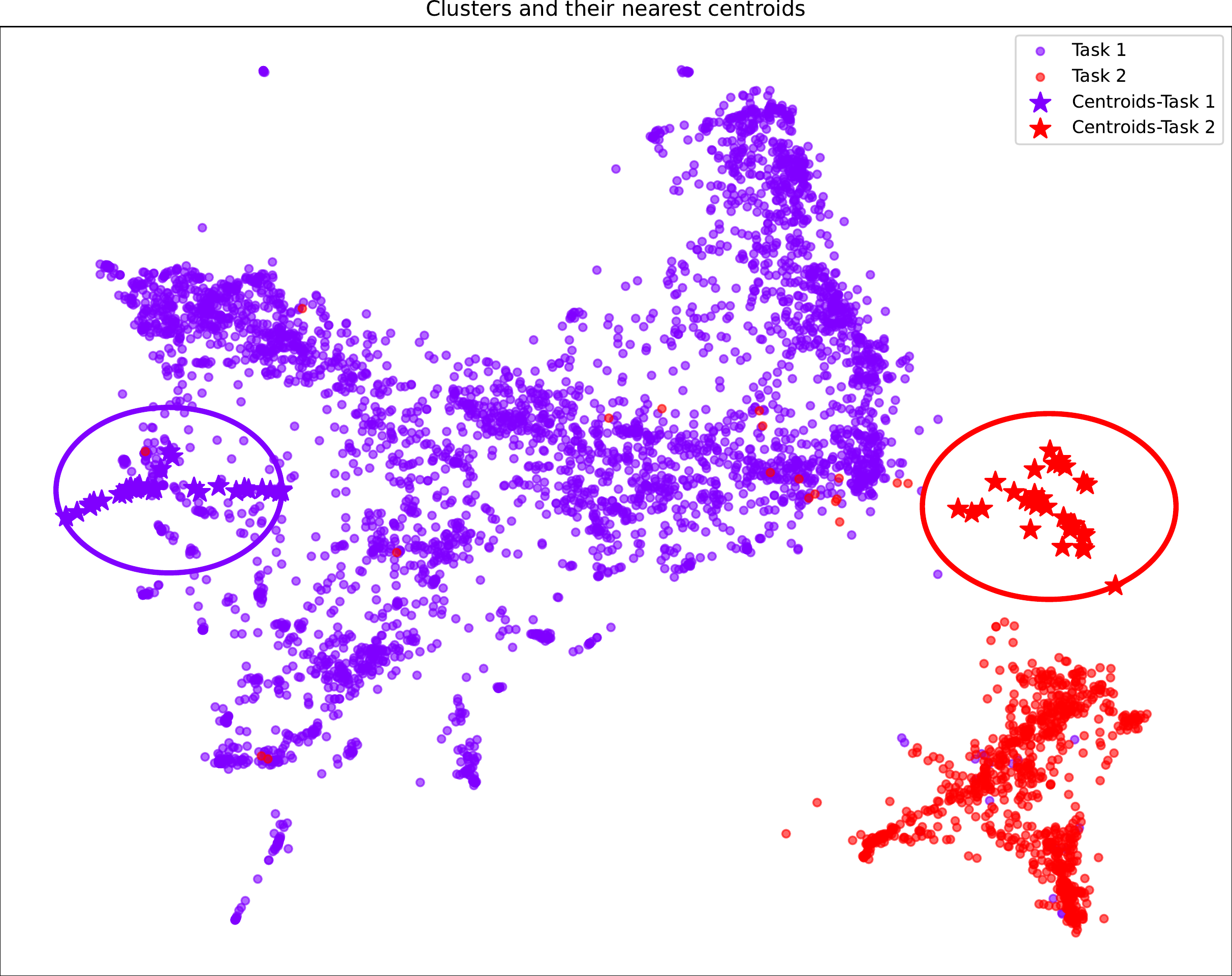}

   \caption{The figure illustrates the outcomes of the density clustering phase, presenting one distinct cluster for each embedding associated with a specific task. For enhanced visualization, two tasks are depicted. Additionally, the nearest neighbors of each centroid for each cluster are displayed above them.}
   \label{fig:clustering_embeddings_2}
\end{figure}

%-------------------------------------------------------------------------
%JG: when describing the experiments, remember to mention the $\epsilon$ and $minPts$ used
% GB: Ok
% JG: Now, I think we better mention the values of $f_{clust}(E; \epsilon, minPts)$ used here
% GB: eps=0.3, minPts=10,
% JG: added
\textbf{Density-based clustering}. Next, we cluster the embeddings $E$ based on their cosine similarity using the DBSCAN density clustering algorithm. Let the resulting clustering labels be $C={c_1,c_2,...,c_m}$, where $c_i$ is the cluster label assigned to the $i$-th embedding $e_i$. Let the clustering function be $f_{clust}(E; \epsilon, minPts)$, where $\epsilon$ is the maximum distance between two points for them to be considered in the same cluster (in our setup, $\epsilon=0.3$) and $minPts$ is the minimum number of points required to form a dense region (in our setup, $minPts=10$). An example of the outcomes of the density-based clustering can be appreciated in Fig.~\ref{fig:clustering_embeddings_2}, which depicts the 2D projection of two clusters of embeddings corresponding to two different tasks. Note that this is just for illustration purposes. The number of dimensions of our embeddings is 512, so one could expect an even clearer distinction between the clusters.

\begin{figure}[t]
  \centering
  % \fbox{\rule{0pt}{2in} \rule{0.9\linewidth}{0pt}}
   \includegraphics[width=1.0\linewidth]{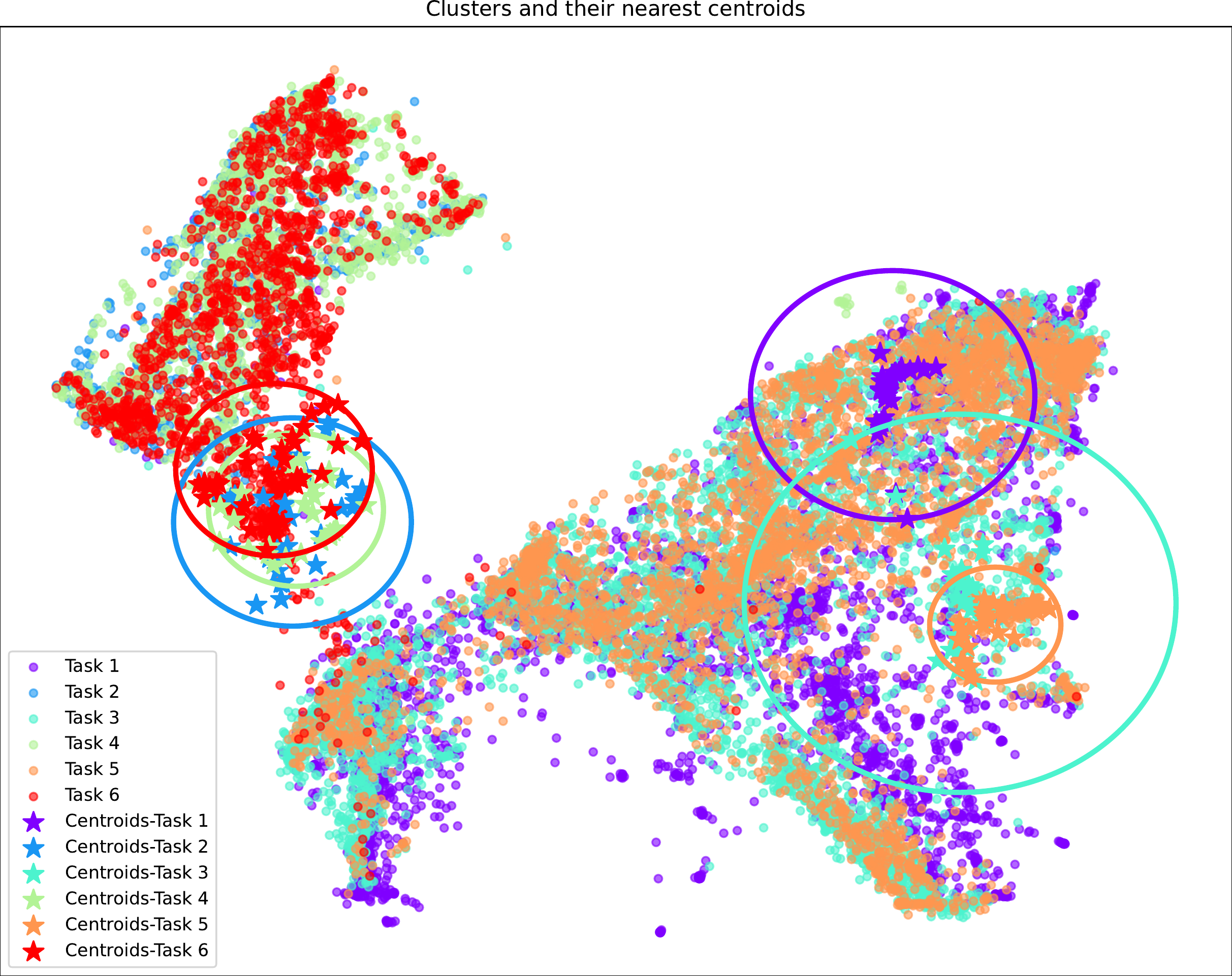}

   \caption{The plot shows the outcomes of the density clustering and the nearest-cluster centroids phases for all the tasks. The nearest neighbors of each centroid for each cluster are displayed on top of them.}
   \label{fig:clustering_embeddings_6}
\end{figure}

%-------------------------------------------------------------------------
%JG: when describing the experiments, remember to mention the $f_{cent}$ function used: should we add a reference for the Nearest Centroid algorithm?
% GB: yes. DONE
% JG: Ok to the reference.
% JG: I mentioned that $f_{cent}$ function uses Euclidean distances
% GB: ok
% JG: I see in the notebook that this is using manhattan distances, so I changed
\textbf{Nearest-cluster centroids}. Then, we obtain the nearest centroids $M={m_1,m_2,...,m_j}$ of the $j$ distinct clusters present in $C$. Each centroid $m_i$ is calculated in the first phase of the Nearest Centroids Algorithm \cite{tibshirani2002diagnosis} from all the embeddings $e_x$ where the cluster label $c_x = i$. We can represent this process as $M=f_{cent}(E,C)$, where $f_{cent}$ is the function that obtains the centroids, in our case, by using Manhattan distances. At this point, we obtain the $k$ nearest neighbors of each centroid $m_i$ from the embeddings $E$ using a nearest-neighbor algorithm such as k-Nearest Neighbors (in our setup, $k=10$). An example of the nearest-cluster centroids can be appreciated in Fig.~\ref{fig:clustering_embeddings_6}, which depicts the nearest neighbors of each centroid (inside the circles) on top of the clusters corresponding to the six different tasks used in our experiments. Recall that this is a 2D projection, and real centroids do not overlap as much.

%-------------------------------------------------------------------------
% JG: Again the 'Nearest Centroid Algorithm'. Is this the same as the previous paragraph?
% GB: yes
% JG: I'm confused: in previous paragraph, 'Nearest Centroid Algorithm' is used to obtain the centroid. Here, it is used to obtain the classifier. This sounds weird
% JG: should we add a reference for the K-Nearest Neighbors algorithm?
% GB: Maybe no because it is common.
% JG: ok
% JG: 'their corresponding task labels for tasks $T_1, T_2, ..., T_{t-1}$': I assume those are the task labels predicted before?
% GB: The ones detected by the detector. By the way, I would refine that part seems not clear
%JG: I clarified a bit the use of the Nearest Centroid Algorithm. I think we still need to define the value of k, right?
% GB: I used k=10
% JG: added
\textbf{Nearest-centroid Incremental classifier}. Finally, we obtain the task classifier $h_t$ for task $T_t$ by running the second phase of the Nearest Centroids Algorithm. Specifically, given the set of $k$ nearest neighbors of each centroid $m_i$ (let it be $N_i$), we train a task classifier $h_t^i$ using the nearest neighbors $N_i$ and their corresponding task labels for tasks $T_1, T_2, ..., T_{t-1}$. The final task classifier $h_t$ for task $T_t$ is obtained by combining the individual classifiers $h_t^i$ using a majority vote. We can represent this process as $h_t=f_{cls}(M_{t^d}, t^d)$, where $f_{cls}$ is the function that obtains the task classifier $h_t$ using the nearest centroids $M_{t^d}$ and $t^d$ being the new task ID detected by the drift detector $R$ (as defined below). Each $M_{t^d}$ is obtained using the training data ${D_i}_{i=1}^{t-1}$ of the previously seen tasks. The task classifier $h_t$ maps an input $x$ to a predicted task label $\hat{t}$, i.e., $h_t(x) = \hat{t}$.

%--------------------------------------------------------------------------
%P: I think the drift is like a trigger to retrain the previous pipeline, so probably we need a paragraph here to say its functionality, maybe we can move the last paragraph here.(This process allows the task classifier to be incrementally updated as new tasks arrive, allowing for effective learning in a domain with a changing task distribution.)
% JG: ok, done

\textbf{Drift detector}. Additionally, in real scenarios, we need a way to decide when to update incrementally the task classifier $h_t$ as new tasks arrive, that is to say, the trigger $t^d$, which will allow for effective learning in a domain with a changing task distribution. In order to detect drift between a pair of tasks $T_t$ and $T_{t'}$ over time, we define a drift function $R$ that measures the dissimilarity between the nearest neighbors at different time points:

\begin{equation}
R(N_t, N_{t'}) = \frac{1}{k} \sum_{s=1}^k d(N_{t[s]}, N_{t'[s]})
\label{eq:drift_detector}
\end{equation}

where $N_t$ and $N_{t'}$ are the sets of $k$ nearest neighbors obtained from the embeddings for tasks $T_t$ and $T_{t'}$, respectively. $d(a, b)$ represents the distance between points $a$ and $b$. This function computes the average distance between the $k$ nearest neighbors in $N_t$ and $N_{t'}$ using the Maximum Mean Discrepancy (MMD) method. A larger value for the drift function indicates a greater difference between the nearest neighbors, suggesting a possible shift in the data distribution, hence, a new task. 
%-------------------------------------------------------------------------

\section{Online algorithm for task-agnostic domain-incremental learning}
\label{sec:pipeline}
%Put here the algorithm that uses all the components. Create a figure that shows this algorithm (use all the previous components). The algorithm that uses all the components is shown in Algorithm \ref{algo:task_classifier}.

\noindent In this section, we present a pipeline algorithm for task-agnostic domain-incremental learning in an online fashion by using the components presented in the previous section (details are shown in Algorithm \ref{algo:task_classifier}).

%P: in the previous section(drift detector), you use Ti and Tj to express the task changing over time, I suppose i meaning before in memory and j meaning new data? But in this section, the new arrival you use Ni. Probably it is better to have the same under-label.
%JG: I changed everywhere with t and t' (generic representation)
% P: need to change as well in figure 1
% JG: true. We'd like to improve the figure to provide more details about the pipeline, but in any case, the notation must be updated
% GB: I did some modifications in the figure
% JG: So did I
For each batch of images arriving as input, our algorithm calculates the nearest-centroid embeddings $N_t$ and then checks whether they exhibit drift with the already known tasks stored in the memory, starting with the most recent. Drift is evaluated through the drift detector introduced before. If the batch drifts regarding all the tasks, that is, it is a new task, we save $N_t$ in memory $\mathcal{M}$, we train incrementally the task classifier $h_t$ using $N_t$ and the new task label $T_t$ with no supervision, and we add a new head corresponding with this new task to the multi-head classifier, which will be used for inference until there is a domain change. Conversely, if we identify in the memory some task that is not drifting from the incoming batch, the classifier $h_t$ is employed to estimate the task ID. As this is expected to be the same as the matching task in the memory, we can add an additional check that will help to identify accuracy issues with the drift detector and/or the task classifier. The task ID is then fed into the multi-head classifier to select the appropriate classifier, which will be used for inference until there is a domain change.

% \begin{algorithm}%[H]
% \SetAlgoLined
% \KwData{$D_i$, $T_i$, $p_{value}=0.05$}
% \KwResult{$h_t$: task classifier, $g_k(e(\mathbf{x}))$: multi-head classifier, $\mathcal{M}$: Memory}
% \eIf{$T_1$}{
% train $g_1({D_1})$;
% get $N_1$ and save into memory $\mathcal{M}$;
% use multi-head classifier $g_1({D_1})$ for inference
% }{
% get $N_i$;
% use $R$ to check if there is drift between $N_i$ and each $N_j$ $\in$ $\mathcal{M}$, for $j<i$

% \eIf{R($N_i$, $N_{i-1}$) $<$ $p_{value}$}{
% save $N_i$ in $\mathcal{M}$;
% train the classifier $h_t$ using $N_i$ and the new task label detected i;
% add a new head to the multi-head classifier;
% }{
% \eIf{there is drift between $N_i$ and another $N_j$ $\in$ $\mathcal{M}$, where $j < i-1$}{
% use the classifier $h_t$ to predict the task;
% use the corresponded head in the multi-head classifier for inference;
% }{
% use the multi-head classifier for inference;
% }
% }
% }
% \caption{Online learning algorithm using the task classifier $h$ and drift detector $R$}
% \label{algo:task_classifier}
% \end{algorithm}$

%P:ok, I see here the i is new and j is in memory, but in drift detector shows R(Ni, Nj), here shows R(Nj, Ni). Better use the same sequence.
%JG: done

\begin{algorithm}[H]
\caption{Online Task-Agnostic algorithm for Domain-Incremental Learning}%\label{alg:alg1}
\begin{algorithmic}
\STATE
\STATE \textbf{function} \textsc{online\_TADIL}$(\mathcal{M}, D_t)$
\STATE \hspace{0.25cm} $N_t \leftarrow$ \textsc{get\_nearest\_centroid\_embeddings}$(D_t)$
\STATE \hspace{0.25cm} \textbf{for} {$N_{t'} \in \mathcal{M}.$\textsc{reversed}()}
    \STATE \hspace{0.5cm} \textbf{if} {(not \textsc{R}($N_t$, $N_{t'}$))}
        \STATE \hspace{0.75cm} \textsc{use} the task classifier $h_{t'}$ to predict the task ID $T_t$ 
        \STATE \hspace{0.75cm} \textbf{if} {($T_t \neq T_{t'}$)} 
            \STATE \hspace{1cm} \textsc{raise warning}
        \STATE \hspace{0.75cm} \textbf{end if}
        \STATE \hspace{0.75cm} \textsc{use} head $g_{t'}({D_{t'}})$ from the classifier for inference
        \STATE \hspace{0.75cm} \textbf{return}
    \STATE \hspace{0.5cm} \textbf{end if}
\STATE \hspace{0.25cm} \textbf{end for}
\STATE \hspace{0.25cm} \textsc{save} $N_t$ into memory $\mathcal{M}$
\STATE \hspace{0.25cm} \textsc{train} incrementally the task classifier $h_t$ using $N_t$ and 
\STATE \hspace{1.3cm} a new task label $T_t$
\STATE \hspace{0.25cm} \textsc{add} a new head $g_t({D_t})$ to the multi-head classifier
\STATE \hspace{0.25cm} \textsc{use} head $g_t({D_t})$ for inference
\STATE \hspace{0.25cm} \textbf{return}
\STATE \textbf{end function}
\end{algorithmic}
\label{algo:task_classifier}
\end{algorithm}

\section{Experimental evaluation}
\label{sec:experiments}
\subsection{Testbed}

% JG: for the project with Lenovo, we will need to run these on Intel CPUs
% GB: Exactly. If we got some time I can modify the code for Intel CPUs
% GB: DONE. Still checking the numbers, not 100% ready
% GB: DONE.
% GB: I think we can make the testbed format better provided that we have more space
% JG: Yes, I agree

\begin{figure}[t]
  \centering
  % \fbox{\rule{0pt}{2in} \rule{0.9\linewidth}{0pt}}
   \includegraphics[width=1.0\linewidth]{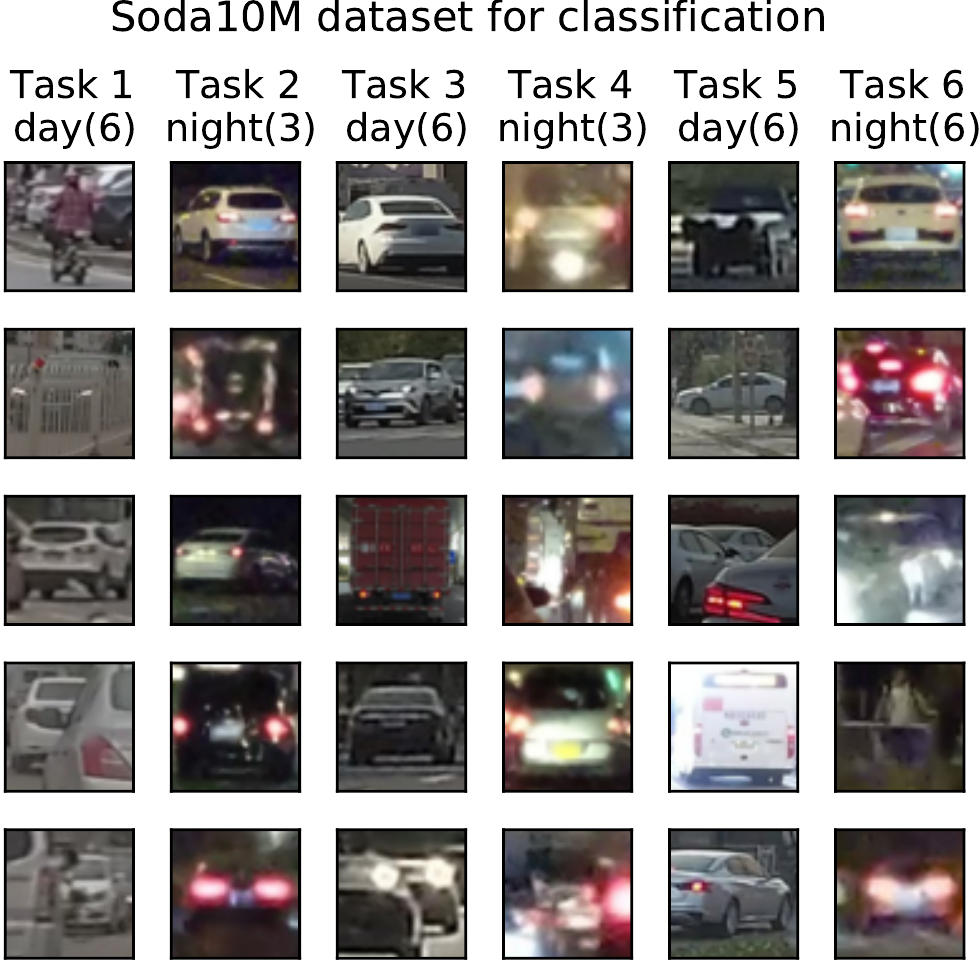}

   \caption{Soda10M for the CLAD-C benchmark. It consists of 6 distinct tasks, each featuring a specific number of classes. For example, Task 1 includes images belonging to 6 classes taken during the day. Similarly, Task 2 includes images belonging to at most 3 classes taken at night. The objective of the multi-head model is to accurately classify images for each individual task.}
   \label{fig:soda10m}
\end{figure}

\noindent The testbed used in the experiments is as follows:
\begin{itemize}
    \item 
    Platform: Ubuntu 22.04 (64 bits).
    \item 
    Hardware: 2x Intel(R) Xeon(R) Platinum 8360Y CPU @ 2.40GHz, 256 GB RAM.
    \item 
    Software: Docker image intel/oneapi-aikit:devel-ubuntu22.04\footnote{https://hub.docker.com/r/intel/oneapi-aikit} (Intel AI Analytics Toolkit), avalanche-lib 0.3.1\footnote{https://avalanche.continualai.org/} (CL library), torch 1.12.0 and torchvision 0.13.0\footnote{https://www.pytorch.org} (DL library), intel-extension-for-pytorch 1.12.100+cpu\footnote{https://github.com/intel/intel-extension-for-pytorch} (Intel acceleration for Pytorch), scikit-learn 1.2.2\footnote{https://scikit-learn.org/stable/} (ML library) and scikit-learn-intelex 2023.0.1\footnote{https://github.com/intel/scikit-learn-intelex} (Intel acceleration for Sklearn).
    \item
    Datasets: SODA10M, which contains 10M unlabeled images and 20k labeled images \cite{han2021soda10m}. We use the labeled images (20,000 1920×1080 color images of 6 different objects) to evaluate our experiments. We created a modified version of the dataset used in the CLAD-C challenge for online classification \cite{verwimp2022clad} (see Fig.~\ref{fig:soda10m}). We split up the 6 tasks into training (80\%) and testing data (20\%) so that we obtain a domain incremental setup for classification. Besides, we tested with different metrics which are aligned to our experiments.
\end{itemize}

% \begin{table}[]
% \caption{Settings for each strategy}
% \label{tab:strategy_settings}
% \addtolength{\tabcolsep}{-1pt}
% \small
% \centering
% \begin{tabular}{l|l} 
% Strategy & Parameters \\
% \hline \hline Shared & $\begin{array}{l}\text { optimizer }=\mathrm{SGD}, \mathrm{lr}=0.01, \text { weight decay }=3 e-5, \\
% \text { momentum }=.99, \\
% \text { nr. blocks }=4 \text { for shifting source, } \mathrm{nr} . \text { blocks }=3 \\
% \text { for transformed }\end{array}$ \\
% \hline EWC & $\lambda=0.4$ \\
% \hline Replay & $\alpha=0.9, \mathrm{kd}=1$ for shifting source, lkd $=0.1$ for transformed \\
% \hline LwF & $T=2$ \\
% \hline
% \end{tabular}
% \end{table}

\subsection{Performance of the drift detector}
\begin{figure}[t]
  \centering
  % \fbox{\rule{0pt}{2in} \rule{0.9\linewidth}{0pt}}
   \includegraphics[width=0.8\linewidth]{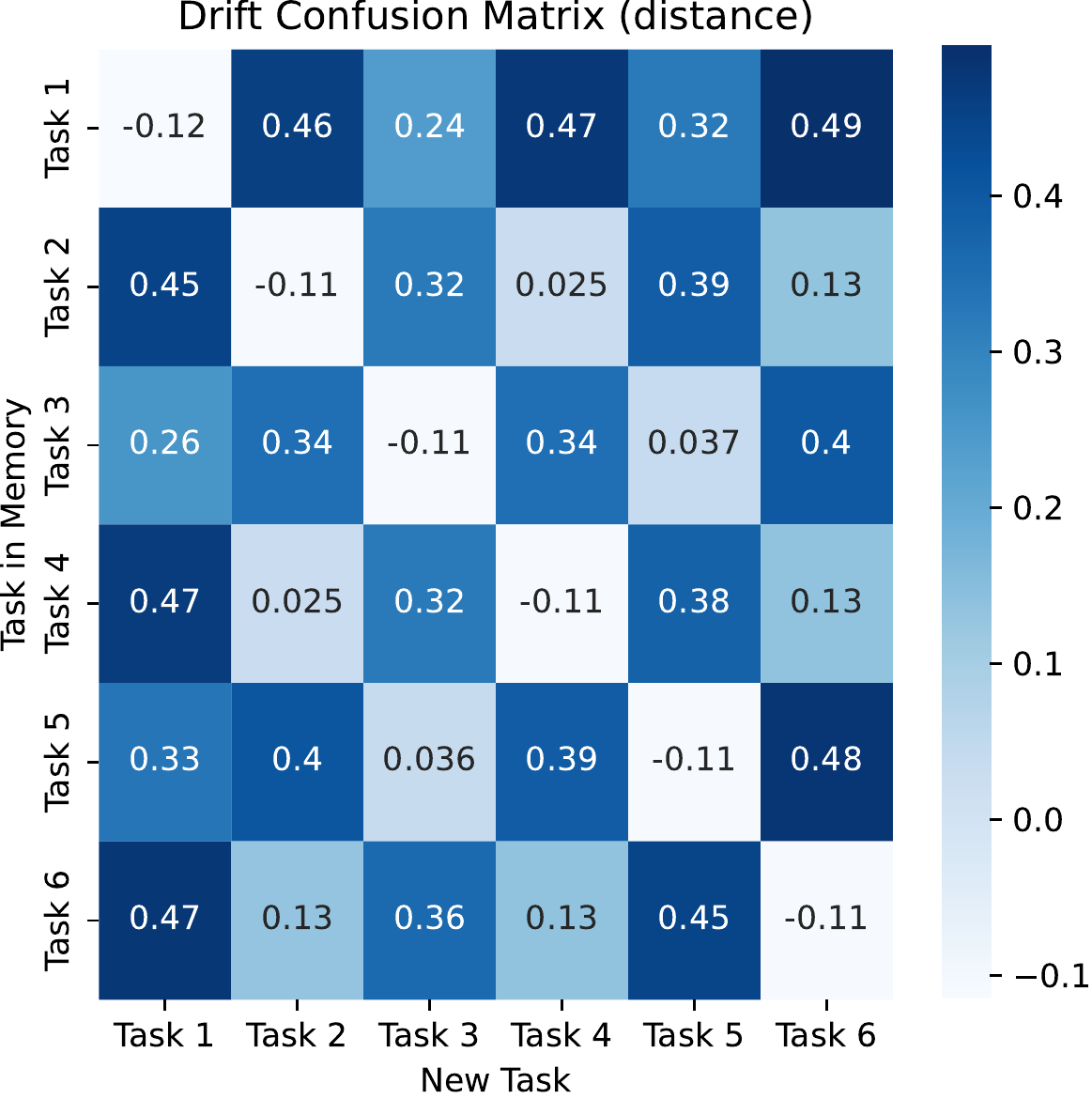}

   \caption{The plot shows the performance of the drift detector. Negative distances indicate that there is no drift, whereas positive ones indicate drift. As shown, the drift detection is 100\% accurate.}
   \label{fig:drift_confusion}
\end{figure}

\noindent In this section, we evaluate the performance of our drift detector component. 
Given a memory component that contains nearest-centroids embeddings from the 6 tasks, we simulate the arrival at inference time of other embeddings from the same tasks. The objective is to see if the drift detector is able to detect the change of boundaries between tasks. Fig.~\ref{fig:drift_confusion} shows the performance of the drift detector by building a confusion matrix to measure the average distance between the $k$ neighbors of each pair of tasks, that is, their dissimilarity. Negative distances indicate there is no drift, whereas positive ones indicate drift. As shown in the matrix, drift is correctly detected every time the new task differs from a former task in the memory (and only in this case).

\subsection{Performance of the task classifier}
\begin{figure}[t]
  \centering
  % \fbox{\rule{0pt}{2in} \rule{0.9\linewidth}{0pt}}
   \includegraphics[width=1.0\linewidth]{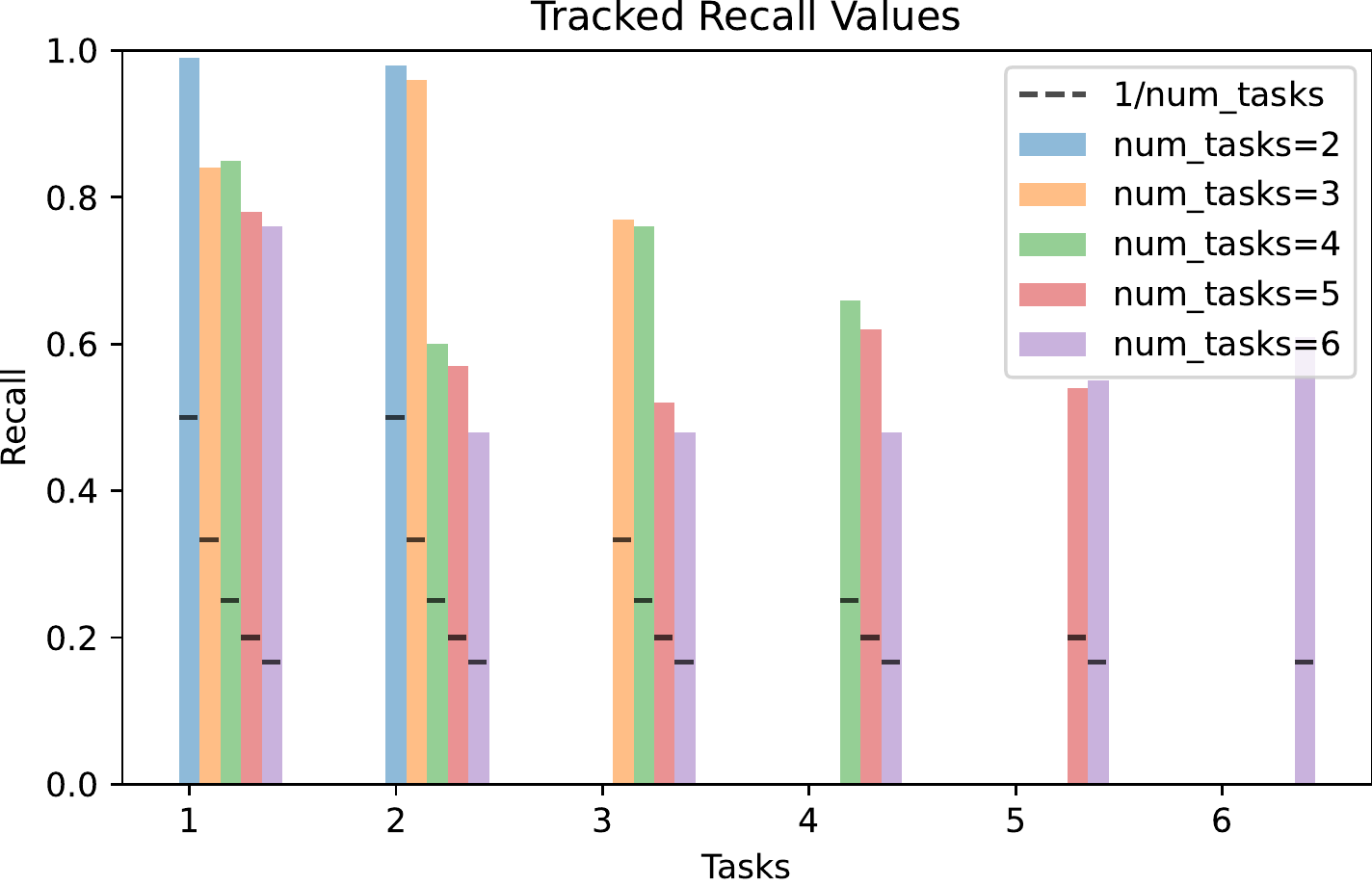}

   \caption{The plot displays the tracked recall values for each task as the number of tasks in the classifier increases from 2 to 6. Each group in the plot represents a task, and within each group, there are bars corresponding to different stages of the classifier. The black lines represent the minimum required recall values, calculated as 1 divided by the number of tasks in the classifier at each stage. This ensures getting the correct prediction for each task.}
   \label{fig:recall}
\end{figure}

\noindent In this section, we evaluate the performance of our task classifier component. 
Fig.~\ref{fig:recall} shows the tracked recall values for each task as the number of tasks in the classifier increases from 2 to 6. Since the most frequently occurring prediction is taken for each batch of images, the task ID prediction achieves 100\% accuracy. This is evident in the plot, as the recall values for all the bars are consistently above the minimum required (the black lines), despite the classifier's accuracy for individual samples is not perfect. This indicates that the classifier performance is sufficient to stay above the minimum required recall values as the number of tasks increases. For example, when the classifier has two tasks (shown in blue in Fig.~\ref{fig:recall}) the black line indicates that the recall of each task in the classifier should be above 0.5 to obtain the perfect prediction by using the mode (i.e., the most frequent value) of the predictions. As observed, both recalls (0.99 and 0.98) fulfill this requirement. 

\subsection{Performance of the CL multi-head models}
\subsubsection{Task-boundary setup}

In this subsection, we compare the performance of some CL strategies when using different approaches to obtain the task ID in a classical CL setup that assumes the existence of task boundaries. In this setup, the data stream is divided into a sequence of tasks, each with a distinct set of classes or concepts. The tasks are presented to the model one by one, and the model has to learn each task without forgetting the previous ones. Being a setup with task boundaries, the drift detection is not necessary. 

The ML model is implemented as a multi-head model that employs an Adam optimizer with a learning rate (lr) of 0.01 and cross-entropy loss as the criterion. The evaluated CL strategies include Elastic Weight Consolidation (EWC), a regularization-based method, Experience Replay, a rehearsal-based method, and Learning without Forgetting (LwF), an architecture-based method. All the strategies were executed using 4 epochs, a batch size of 200, and the same optimizer and criterion, with the remaining parameters set to their default values as defined in the Avalanche library.

The multi-head model is also given a task ID or a task label for each task, which indicates which task it is currently learning. This allows the model to switch between different output heads or parameters for different tasks. The model is evaluated on its accuracy on all known tasks after learning each new task. We compare three different approaches to supply the task ID, namely the \textit{ground-truth} approach, where the task ID is known in advance, our approach (\textit{TADIL}), where the CL strategies use the task ID supplied by our task classifier, and the \textit{normal} approach, in which the strategies do not receive the task ID.

% GB: These figures will be updated
% JG: Given that we do not have space constraints (for the moment), I expanded the figures to multicolumn
% GB: ok
\begin{figure*}[t]
  \centering
  % \fbox{\rule{0pt}{2in} \rule{0.9\linewidth}{0pt}}
   \includegraphics[width=0.9\linewidth]{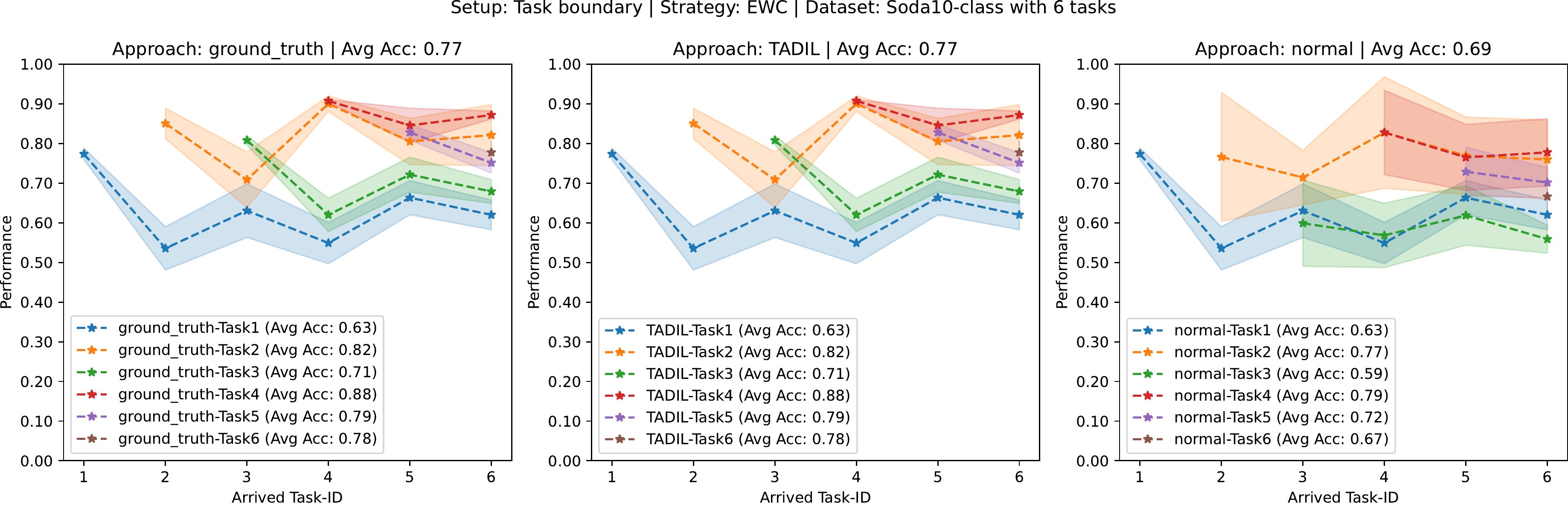}

   \caption{Comparison using the EWC strategy. This figure shows how this strategy behaves with task boundaries.}
   \label{fig:ewc}
%\end{figure*}
\vspace*{\floatsep}
%\begin{figure*}[t]
%  \centering
  % \fbox{\rule{0pt}{2in} \rule{0.9\linewidth}{0pt}}
   \includegraphics[width=0.9\linewidth]{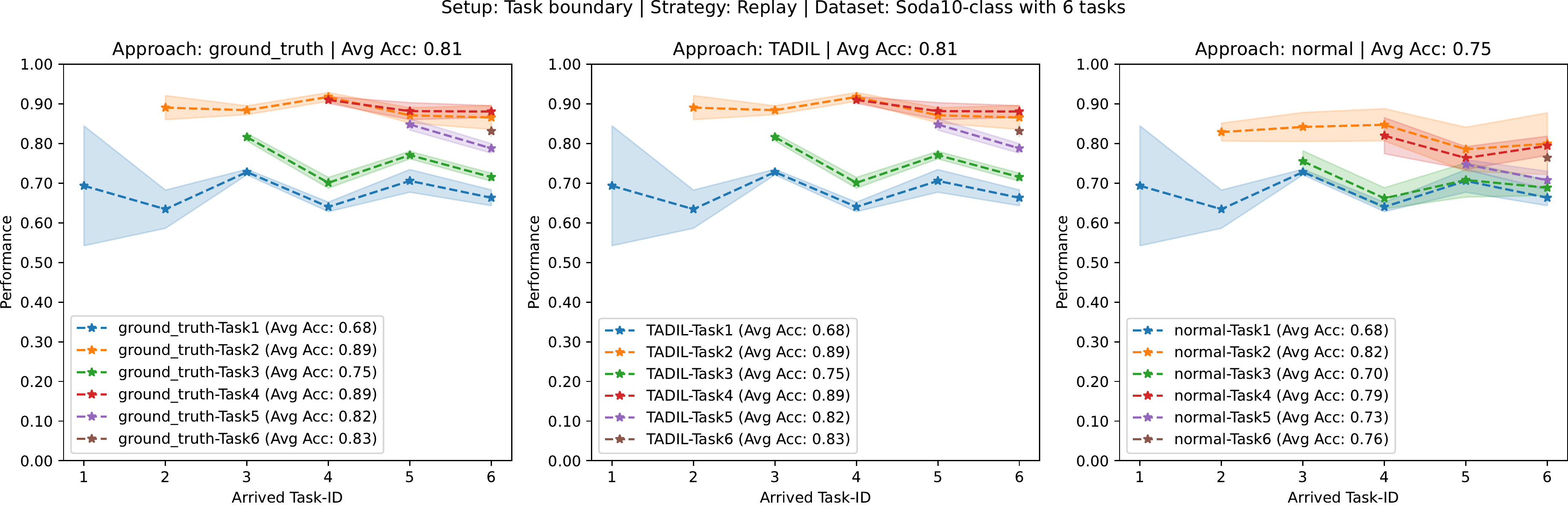}

   \caption{Comparison using the Experience Replay strategy. This figure shows how this strategy behaves with task boundaries.}
   \label{fig:replay}
%\end{figure*}
\vspace*{\floatsep}
% \begin{figure}[t]
%   \centering
%   % \fbox{\rule{0pt}{2in} \rule{0.9\linewidth}{0pt}}
%    \includegraphics[width=1.0\linewidth]{GEM.pdf}

%    \caption{GEM (Gradient Episodic Memory) strategy.}
%    \label{fig:gem}
% \end{figure}

%\begin{figure*}[t]
%  \centering
  % \fbox{\rule{0pt}{2in} \rule{0.9\linewidth}{0pt}}
   \includegraphics[width=0.9\linewidth]{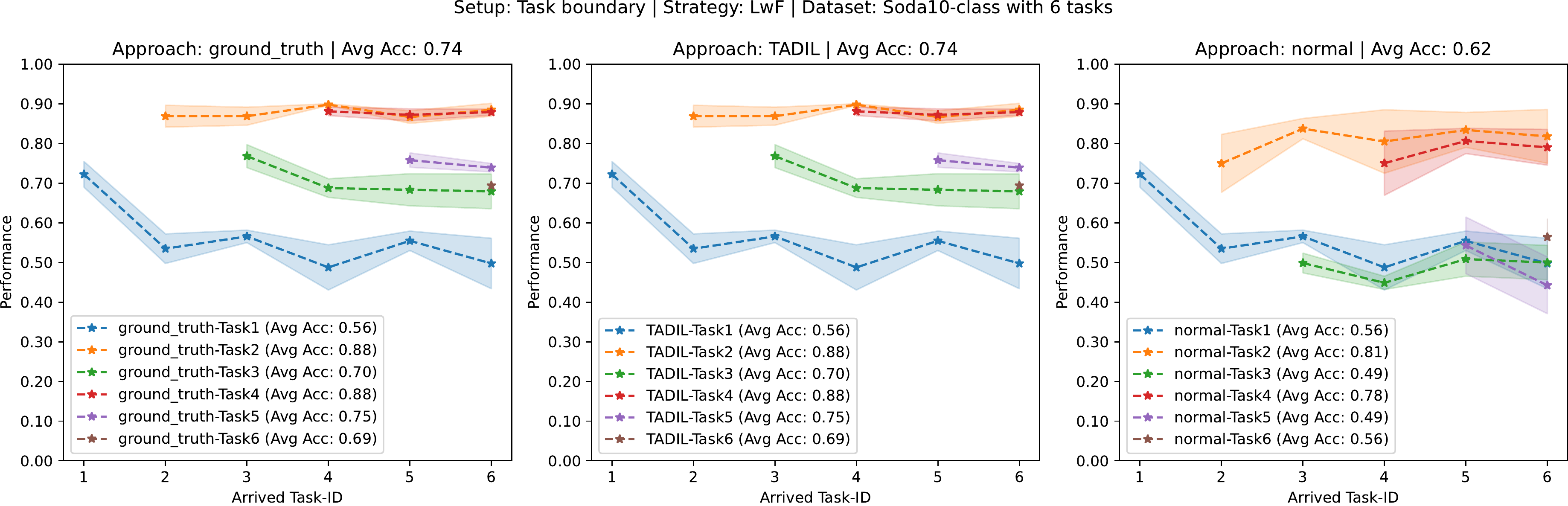}

   \caption{Comparison using the LwF strategy. This figure shows how this strategy behaves with task boundaries.}
   \label{fig:lwf}
\end{figure*}

Fig. \ref{fig:ewc}, \ref{fig:replay}, and \ref{fig:lwf} present the accuracy of the EWC, Replay, and LwF strategies, respectively, through three subplots. The first one illustrates the \textit{ground-truth} scenario, the second subplot displays our approach (\textit{TADIL}), and the third subplot depicts the \textit{normal} scenario. For each strategy and scenario, we did 4 executions with different seeds. Dashed lines display the average values, whereas the shaded areas show the standard deviations. Each line shows the performance of a given task while new tasks are arriving. For instance, the blue line corresponds to the performance of Task 1, which is repeatedly evaluated upon the incremental arrival of tasks 2, 3, 4, 5, and 6 (i.e., Task 1, Task 1 + Task 2, Task 1 + Task 2 + Task 3, \dots, Task 1 + Task 2 + \dots + Task 6). Similarly, the yellow line corresponds to the performance of Task 2 and is repeatedly evaluated upon the incremental arrival of tasks 3, 4, 5, and 6. Finally, Task 6 is only a dot (in brown color) as there are no more new tasks from that point on. %Table \ref{tab:all_accuracies} summarizes the average accuracy for each scenario.

The plots show that our proposed method outperforms the \textit{normal} approach oblivious of the task ID and is on par with the \textit{ground-truth} approach in terms of acquiring the correct task ID for the multi-head model. This can be attributed to our task classifier's delivery of the most frequently occurring prediction. Consequently, this enables 100\% accuracy on the task ID prediction.%, even if the classifier's accuracy for individual samples is not perfect. 

%JG rewrite next paragraph
% The impact of knowing the task ID is larger for EWC compared to Experience Replay and LwF when using a multi-head classifier as a base model, because EWC struggles to assign the correct importance to weights without knowing which task the new data belongs to. On the other hand, Experience Replay can still access stored experiences and LwF can use the teacher model to maintain knowledge about previous tasks, making them more robust to the absence of the task ID information. 
In all the CL strategies, when the multi-head model is faced with a new task, it can use the task ID as a form of reference to distinguish between tasks and to decide when to apply previously learned knowledge. Consequently, knowing the task ID improves the performance for all the strategies, although the impact is less significant for EWC and Experience Replay in comparison with LwF. This occurs because the LwF strategy has a stronger dependency on the task ID as it directly uses the soft targets generated from the old network for a given task when it receives an already seen task ID. Contrariwise, EWC and Experience Replay focus more on balancing the weights and experiences from past tasks with the new ones.

Whereas the three CL strategies are designed to alleviate the forgetting problem, this can still happen when the new task is very different from the known ones, especially with EWC. For instance, the performance of Task 3, which includes images taken during the day, gets worse upon the arrival of Task 4, which includes images taken during the night. Notably, the performance can improve again when a more similar task arrives. For instance, the performance of Task 3 improves upon the arrival of Task 5.

\subsubsection{Task-agnostic setup}
% The more realistic scenarios should be located here. It includes the drift detector and the online algorithm.
% JG: we need to explain in more detail the results in the plots, especially regarding the 'forgetting' problem
% JG: Done
The previous task-boundary setup has been widely used in many CL benchmarks and methods. However, it also has some limitations that make it less realistic and applicable to real-world scenarios. First, it requires explicit signals or labels to indicate when a task boundary occurs and which task it belongs to. This may not be available or feasible in many cases where the data stream is continuous and heterogeneous. And second, it assumes that each task has a clear and fixed set of classes or concepts that do not overlap or change over time. This may not hold in many cases where the data stream may exhibit gradual or abrupt changes in its underlying distribution or concept over time. 

\begin{figure*}[t]
  \centering
  % \fbox{\rule{0pt}{2in} \rule{0.9\linewidth}{0pt}}
   \includegraphics[width=0.9\linewidth]{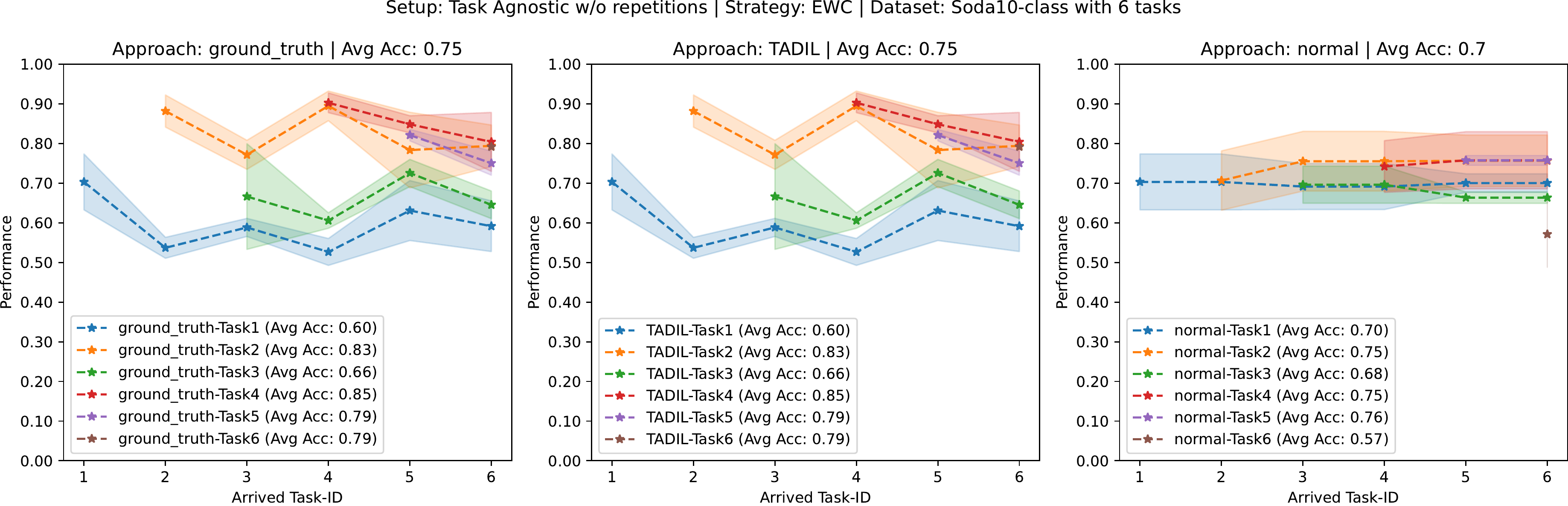}

   \caption{Comparison using the EWC strategy. This figure shows how this strategy behaves with no task boundaries.}
   \label{fig:agnostic_ewc}
%\end{figure*}
\vspace*{\floatsep}
%\begin{figure*}[t]
%  \centering
  % \fbox{\rule{0pt}{2in} \rule{0.9\linewidth}{0pt}}
   \includegraphics[width=0.9\linewidth]{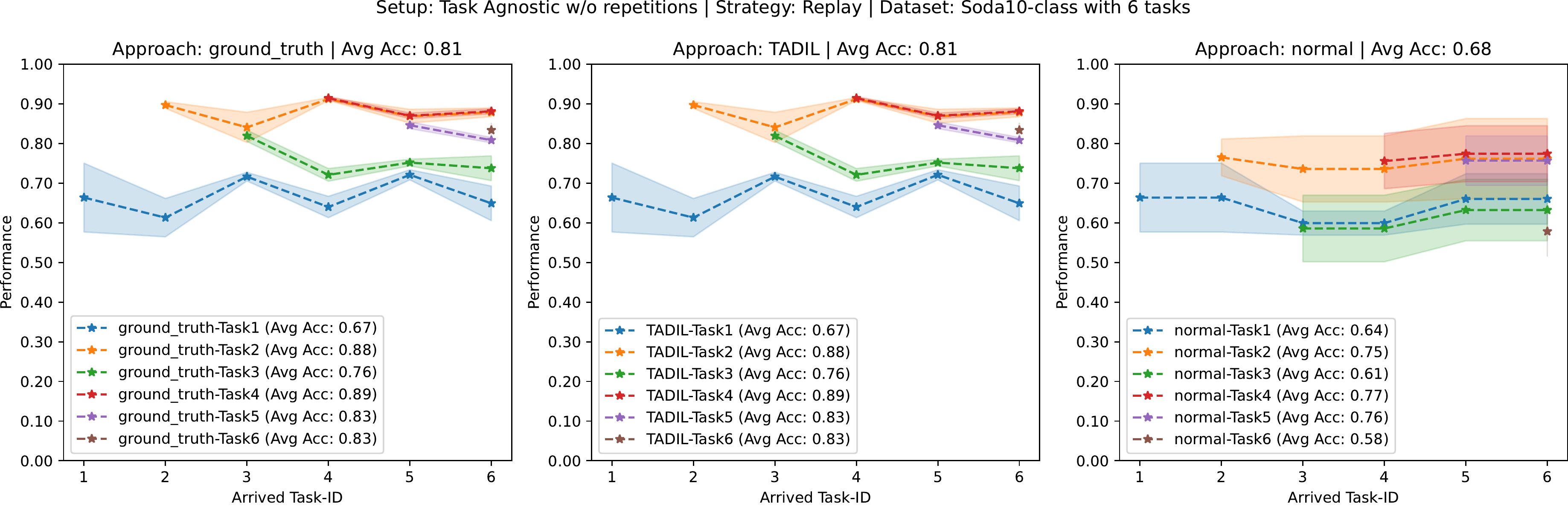}

   \caption{Comparison using the Experience Replay strategy. This figure shows how this strategy behaves with no task boundaries.}
   \label{fig:agnostic_replay}
%\end{figure*}
\vspace*{\floatsep}
%\begin{figure*}[t]
%  \centering
  % \fbox{\rule{0pt}{2in} \rule{0.9\linewidth}{0pt}}
   \includegraphics[width=0.9\linewidth]{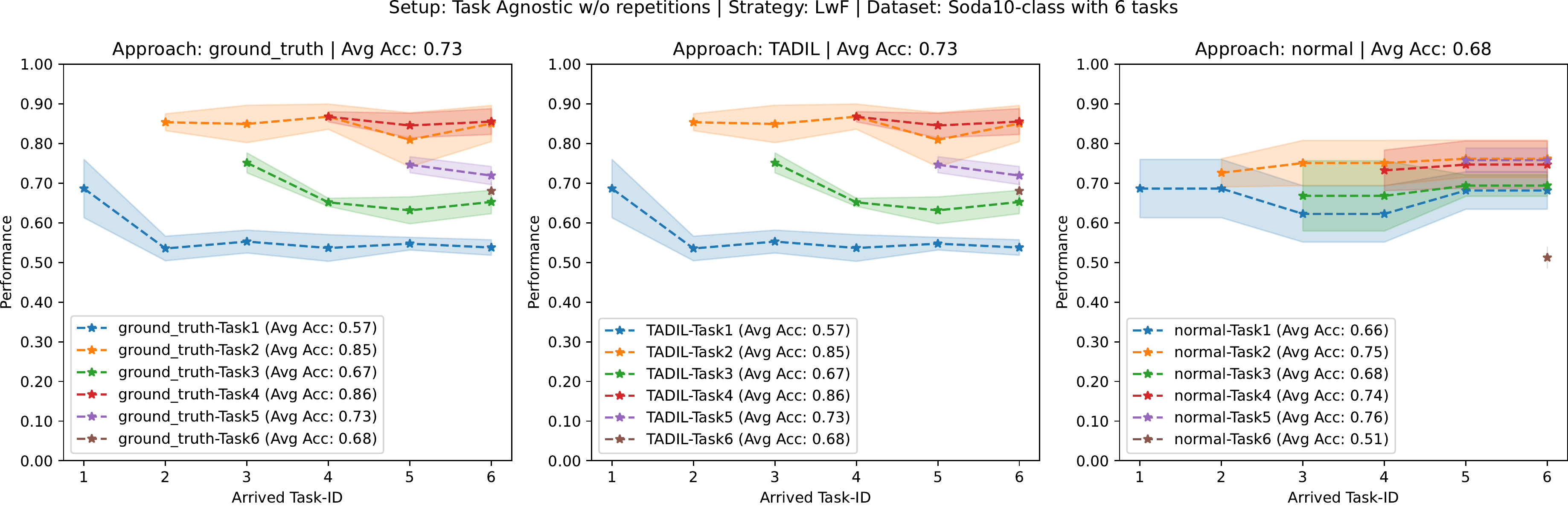}

   \caption{Comparison using the LwF strategy. This figure shows how this strategy behaves with no task boundaries.}
   \label{fig:agnostic_lwf}
\end{figure*}

% JG: Figure moved here to ensure a nicer placement in the PDF
\begin{figure*}[t]
  \centering
  % \fbox{\rule{0pt}{2in} \rule{0.9\linewidth}{0pt}}
   \includegraphics[width=0.9\linewidth]{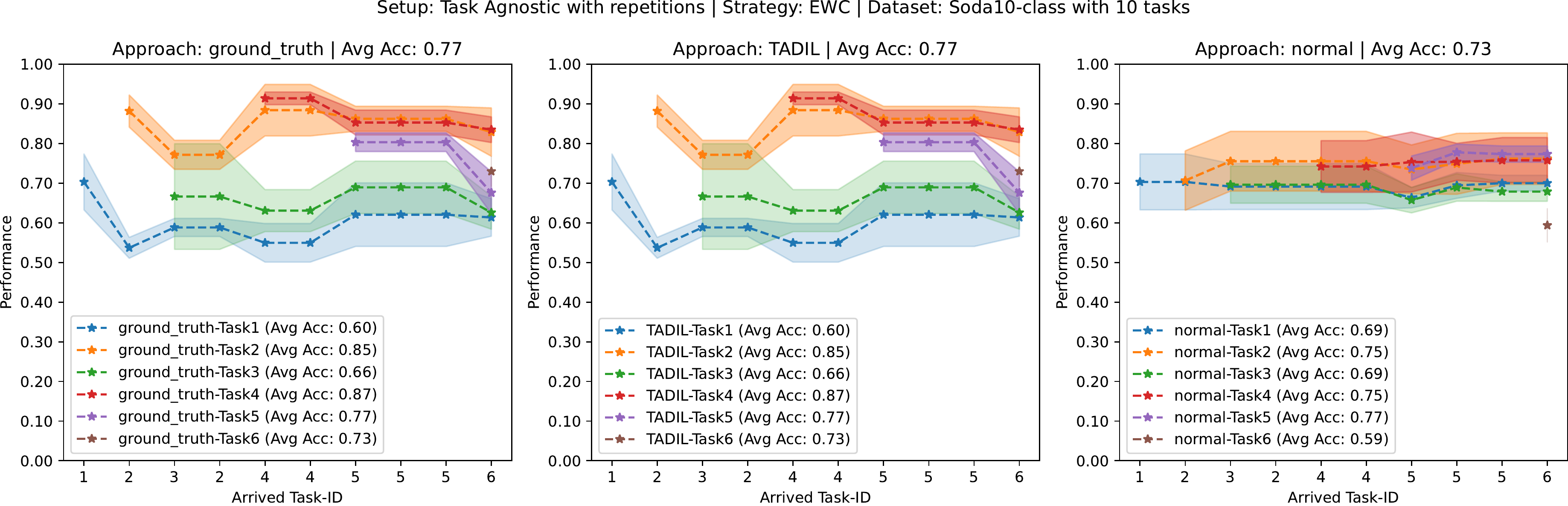}

   \caption{Comparison using the EWC strategy. This figure shows how this strategy behaves with no boundaries and repeated tasks.}
   \label{fig:agnostic_online_ewc}
%\end{figure*}
\vspace*{\floatsep}
%\begin{figure*}[t]
%  \centering
  % \fbox{\rule{0pt}{2in} \rule{0.9\linewidth}{0pt}}
   \includegraphics[width=0.9\linewidth]{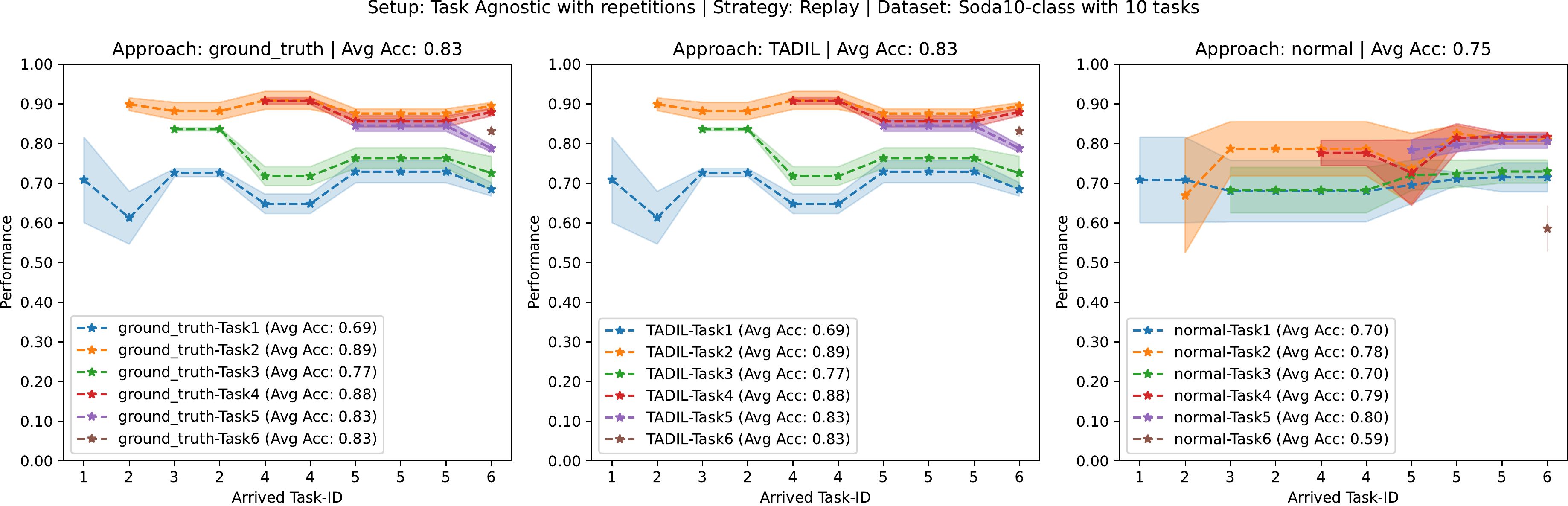}

   \caption{Comparison using the Experience Replay strategy. This figure shows how this strategy behaves with no boundaries and repeated tasks.}
   \label{fig:agnostic_online_replay}
%\end{figure*}
\vspace*{\floatsep}
%\begin{figure*}[t]
%  \centering
  % \fbox{\rule{0pt}{2in} \rule{0.9\linewidth}{0pt}}
   \includegraphics[width=0.9\linewidth]{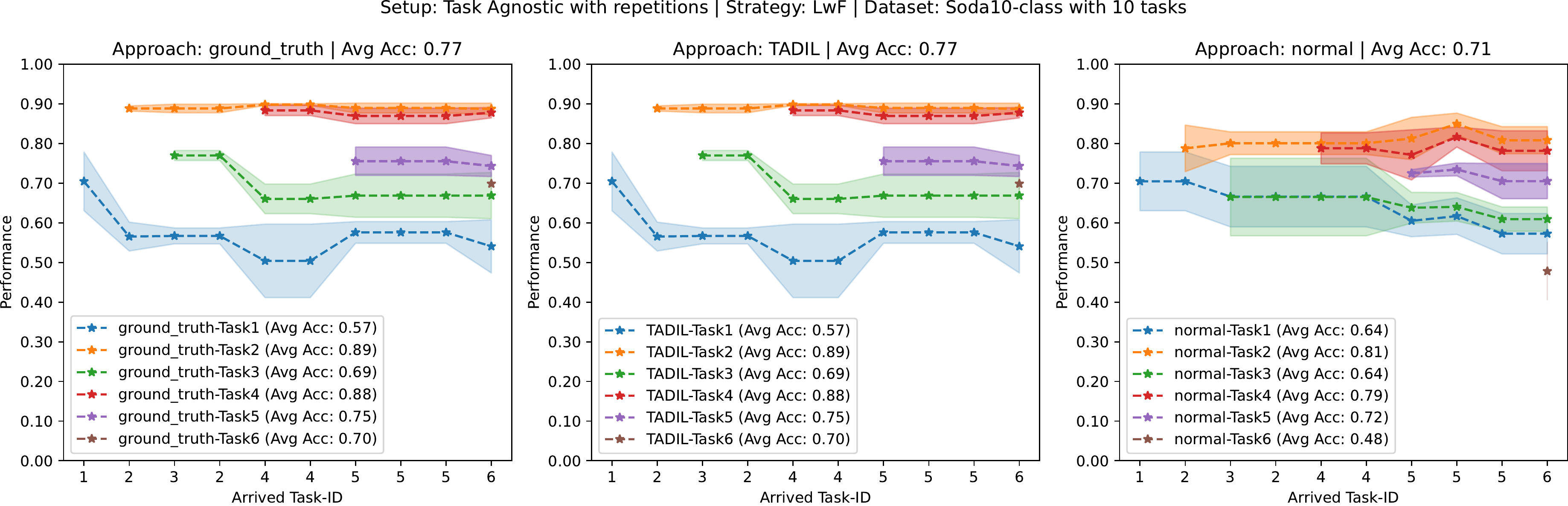}

   \caption{Comparison using the LwF strategy. This figure shows how this strategy behaves with no boundaries and repeated tasks.}
   \label{fig:agnostic_online_lwf}
\end{figure*}

Consequently, in this subsection, we evaluate the former three CL strategies in a task-agnostic scenario to show how our approach is able to detect and identify new tasks correctly. We use the same multi-head model as in the previous experiments, the same dataset sequence (tasks from 1 to 6 arriving in order), and the same CL strategies. However, to detect the appearance of a new task, we use our drift detector component defined in Equation \ref{eq:drift_detector}. Therefore, each of the strategies (EWC, Replay, LwF) will trigger training only when there is a drift caused by the appearance of a new batch of data (i.e., a new task) during the inference stage, which is a more realistic scenario. 

In order to ensure a fair comparison, the \textit{normal} approach which is oblivious of the task ID should also undergo some kind of retraining. In this case, the trigger is different. In essence, this approach is subject to retraining on the arrival of tasks associated with daytime (Task 1, Task 3, and Task 5), even in the absence of explicit task IDs. This setup offers a more equitable point of comparison relative to a scenario where the approach is exclusively trained on the initial task (Task 1). The rationale behind this design choice is worth discussing. If a strategy does not have access to the task IDs during the inference phase, it might be beneficial to allow it to retrain its model periodically. In the context of our investigation, this retraining phase is scheduled to occur at the start of each day. 

Fig. \ref{fig:agnostic_ewc}, \ref{fig:agnostic_replay}, and \ref{fig:agnostic_lwf}  present the accuracy of EWC, Replay, and LwF  strategies, respectively, in this task-agnostic scenario. Again, our proposed method (\textit{TADIL}) outperforms the accuracy of the \textit{normal} method oblivious of the task ID (especially in tasks with nighttime images) and is on par with the \textit{ground-truth} approach in terms of acquiring the correct task ID for the multi-head model. This confirms that providing the task ID improves the accuracy of the multi-head models (as we also showed in the previous experiments), but also that our drift detector is able to identify new tasks correctly. 
% Notably, the performance for the three CL strategies is slightly lower in this task-agnostic scenario, as the drift detector might need some time to realize that there is a domain change. This time is directly related with the batch size.

% \begin{table}[!t]
% \caption{An Example of a Table\label{tab:table1}}
% \centering
% \begin{tabular}{|c||c|}
% \hline
% One & Two\\
% \hline
% Three & Four\\
% \hline
% \end{tabular}
% \end{table}

% JG: Table moved here to ensure a nicer placement in the PDF
\begin{table*}
\caption{Average accuracy and standard errors for each task, CL strategy, and domain-IL approach.}
\label{tab:all_accuracies}
\addtolength{\tabcolsep}{-1.5pt}
\small
\centering

\begin{tabular}{cc|ccc|ccc|ccccccc}
% \toprule
\hline
       &        & \multicolumn{3}{c}{\textbf{Task agnostic without repetitions}} & \multicolumn{3}{c}{\textbf{Task agnostic with repetitions}} & 
\multicolumn{3}{c}{\textbf{With task boundaries}} \\
\hline
 Strategy      &    Task    &           G. Truth & \textbf{TADIL} & Normal &             G. Truth & 
\textbf{TADIL} & Normal &  
G. Truth & \textbf{TADIL} & Normal \\
\hline
% \midrule

EWC  &  1  &              0.60 ±0.05  & \textbf{0.60 ±0.05} &   0.70 ±0.05  &                0.60 ±0.05  & 
\textbf{0.60 ±0.05} &   0.69 ±0.05  &     0.63 ±0.04  & \textbf{0.63 ±0.04} &   0.63 ±0.04 \\
        &  2  &              0.83 ±0.05  & \textbf{0.83 ±0.05} &   0.75 ±0.07  &                0.85 ±0.04  & 
\textbf{0.85 ±0.04} &   0.75 ±0.07  &     0.82 ±0.05  & \textbf{0.82 ±0.05} &   0.77 ±0.11 \\
        &  3  &              0.66 ±0.06  & \textbf{0.66 ±0.06} &   0.68 ±0.03  &                0.66 ±0.08  & 
\textbf{0.66 ±0.08} &   0.69 ±0.04  &     0.71 ±0.03  & \textbf{0.71 ±0.03} &   0.59 ±0.08 \\
        &  4  &              0.85 ±0.04  & \textbf{0.85 ±0.04} &   0.75 ±0.07  &                0.87 ±0.03  & 
\textbf{0.87 ±0.03} &   0.75 ±0.06  &     0.88 ±0.02  & \textbf{0.88 ±0.02} &   0.79 ±0.09 \\
        &  5  &              0.79 ±0.02  & \textbf{0.79 ±0.02} &   0.76 ±0.01  &                0.77 ±0.03  & 
\textbf{0.77 ±0.03} &   0.77 ±0.02  &     0.79 ±0.02  & \textbf{0.79 ±0.02} &   0.72 ±0.05 \\
        &  6  &              0.79 ±0.01  & \textbf{0.79 ±0.01} &   0.57 ±0.08  &                0.73 ±0.04  & 
\textbf{0.73 ±0.04} &   0.59 ±0.04  &     0.78 ±0.03  & \textbf{0.78 ±0.03} &   0.67 ±0.05 \\
\hline
LwF  &  1  &              0.57 ±0.03  & \textbf{0.57 ±0.03} &   0.66 ±0.06  &                0.57 ±0.05  & 
\textbf{0.57 ±0.05} &   0.64 ±0.06  &     0.56 ±0.04  & \textbf{0.56 ±0.04} &   0.56 ±0.04 \\
        &  2  &              0.85 ±0.04  & \textbf{0.85 ±0.04} &   0.75 ±0.05  &                0.89 ±0.01  & 
\textbf{0.89 ±0.01} &   0.81 ±0.04  &     0.88 ±0.02  & \textbf{0.88 ±0.02} &   0.81 ±0.06 \\
        &  3  &              0.67 ±0.02  & \textbf{0.67 ±0.02} &   0.68 ±0.06  &                0.69 ±0.04  & 
\textbf{0.69 ±0.04} &   0.64 ±0.07  &     0.70 ±0.03  & \textbf{0.70 ±0.03} &   0.49 ±0.03 \\
        &  4  &              0.86 ±0.03  & \textbf{0.86 ±0.03} &   0.74 ±0.06  &                0.88 ±0.02  & 
\textbf{0.88 ±0.02} &   0.79 ±0.04  &     0.88 ±0.01  & \textbf{0.88 ±0.01} &   0.78 ±0.05 \\
        &  5  &              0.73 ±0.02  & \textbf{0.73 ±0.02} &   0.76 ±0.03  &                0.75 ±0.03  & 
\textbf{0.75 ±0.03} &   0.72 ±0.03  &     0.75 ±0.01  & \textbf{0.75 ±0.01} &   0.49 ±0.07 \\
        &  6  &              0.68 ±0.02  & \textbf{0.68 ±0.02} &   0.51 ±0.03  &                0.70 ±0.01  & 
\textbf{0.70 ±0.01} &   0.48 ±0.07  &     0.69 ±0.01  & \textbf{0.69 ±0.01} &   0.56 ±0.05 \\
\hline
Replay  &  1  &              0.67 ±0.04  & \textbf{0.67 ±0.04} &   0.64 ±0.06  &                0.69 ±0.03  & 
\textbf{0.69 ±0.03} &   0.70 ±0.07  &     0.68 ±0.04  & \textbf{0.68 ±0.04} &   0.68 ±0.04 \\
        &  2  &              0.88 ±0.02  & \textbf{0.88 ±0.02} &   0.75 ±0.08  &                0.89 ±0.02  & 
\textbf{0.89 ±0.02} &   0.78 ±0.06  &     0.89 ±0.02  & \textbf{0.89 ±0.02} &   0.82 ±0.05 \\
        &  3  &              0.76 ±0.02  & \textbf{0.76 ±0.02} &   0.61 ±0.08  &                0.77 ±0.02  & 
\textbf{0.77 ±0.02} &   0.70 ±0.05  &     0.75 ±0.01  & \textbf{0.75 ±0.01} &   0.70 ±0.03 \\
        &  4  &              0.89 ±0.01  & \textbf{0.89 ±0.01} &   0.77 ±0.07  &                0.88 ±0.01  & 
\textbf{0.88 ±0.01} &   0.79 ±0.03  &     0.89 ±0.01  & \textbf{0.89 ±0.01} &   0.79 ±0.03 \\
        &  5  &              0.83 ±0.01  & \textbf{0.83 ±0.01} &   0.76 ±0.06  &                0.83 ±0.01  & 
\textbf{0.83 ±0.01} &   0.80 ±0.02  &     0.82 ±0.01  & \textbf{0.82 ±0.01} &   0.73 ±0.02 \\
        &  6  &              0.83 ±0.02  & \textbf{0.83 ±0.02} &   0.58 ±0.06  &                0.83 ±0.01  & 
\textbf{0.83 ±0.01} &   0.59 ±0.06  &     0.83 ±0.01  & \textbf{0.83 ±0.01} &   0.76 ±0.05 \\
% \bottomrule
\end{tabular}

\end{table*}

For our \textit{TADIL} approach, we observe that there are not significant discrepancies between the outcomes of the task-boundary and the task-agnostic scenarios. This is due to the good accuracy of our drift detector, which ensures a timely retraining of the multi-head model. Conversely, for the \textit{normal} approach, the performance difference between task-boundary and task-agnostic scenarios has become evident. In this approach, a fresh model is only trained at the beginning of each day, making tasks with daytime images such as Task 1, Task 3, and Task 5 less susceptible to forgetting, because they are not concurrently trained with tasks with nighttime images. This different behaviour was particularly noticeable with Task 1 and differs from our \textit{TADIL} approach, where the performance of a task with daytime images (Task 1) can decline after the arrival of a task with nighttime images (Task 2). However, the performance improves again to some extent when a more similar task, such as Task 3, is introduced.

This situation presents an interesting trade-off when deciding how frequently to update a model: its soundness with the new domains versus the degree of task forgetting that results from the model updates. Nonetheless, despite this trade-off, \textit{TADIL} consistently outperforms the \textit{normal} approach across all the scenarios. This superiority is particularly pronounced in the Replay strategy, where task forgetting was observed to be the lowest among all tasks in comparison with the other strategies.

\subsubsection{Task-agnostic setup with tasks repetitions}
In this subsection, we evaluate the former three CL strategies in a task-agnostic scenario where tasks can repeat over time. This allows mimicking the behavior of many real-world scenarios. For example, the model transitions from processing clear images of the city center during the afternoon to darker images taken at midnight in the countryside, before receiving clear images from the city center again at sunrise. We devised a custom task sequence specifically tailored to test this scenario.

Fig. \ref{fig:agnostic_online_ewc}, \ref{fig:agnostic_online_replay}, and \ref{fig:agnostic_online_lwf} present the accuracy of EWC, Replay, and LwF strategies, respectively, in this task-agnostic scenario with tasks repetitions when the model receives the sequence defined as [1, 2, 3, 2, 4, 4, 5, 5, 5, 6] (where each number represents a particular task). Again, the performance of our proposed method (\textit{TADIL}) is on par with the \textit{ground-truth} approach and is significantly better than the accuracy of the \textit{normal} method oblivious of the task ID, mainly in tasks with nighttime images. 
% update next paragraph
% Despite the fact that the numerical differences aren't too divergent from previous scenarios, a notable distinction appears in the LwF strategy, especially with Task 5, which follows two Task 4 instances (which are night tasks hence no training), as shown in Table \ref{tab:all_accuracies}. The accuracy gap between the "Normal" approach across all scenarios is significant, likely due to the influence of repeated tasks which may intensify with longer task sequences. 

In comparison with the preceding experiment, the performance for our \textit{TADIL} approach remains consistent across all strategies and tasks. Conversely, the \textit{normal} approach is impacted to a higher extent by the specific retraining intervals, which could range from long periods without retraining the model (when several tasks with nighttime images arrive consecutively, as happens with Tasks 2, 4, and 4 in our sequence) to successive intervals of retraining in a row (on the arrival of repeated tasks with daytime images, as happens with Task 5 in our sequence). %For instance, with the Replay strategy, the performance of Task 4 deteriorates more when Task 5 arrives, compared to the previous experiment. %This is possibly due to an overtraining of task 5 that causes it to forget task 4 a bit.
% In comparison to the preceding experiment, the alterations implemented were relatively minor. However, these changes produced some differences in the experimental outcomes. Across all tested scenarios, the performance in the repetitive task, Task 4, consistently deteriorated when Task 5 was introduced. This degradation may be attributed to a prolonged cessation of the model's training when Task 5 is encountered, potentially leading to underfitting at this juncture. This pattern is suggestive of a possible interaction between Task 4 and Task 5 that warrants further investigation to elucidate its underpinning mechanisms.

This continual training associated with repeated tasks denotes also a notable distinction between this experiment and the prior one. Our results indicate that such training can become redundant and inefficient in the absence of task detection, and may even lead to model overfitting. For instance, see how performance of Task 5 with the LwF strategy in the \textit{normal} approach deteriorates after the third consecutive retraining with the same task. This redundancy is particularly critical in contexts where computational resources are constrained. This finding underscores the importance of implementing task detection mechanisms in environments with hardware limitations, as this could significantly optimize the continual training process.

% Finally, the same forgetting behaviour appears than in the previous scenarios. However, repetitive tasks might worsen this forgetting if not drift detection is applied.  

% [explanation]. That is, our approach prevents this forgetting by using the drift detector and the task classifier to check if there is a need to train the model or the information in the memory can be used.
% JG Add missing explanation
% GB: added more info, but still in progress.

Table \ref{tab:all_accuracies} summarizes the accuracy of three approaches to get the task ID when applied to each CL strategy on different scenarios. 
We see that our approach continues to enhance the performance of the strategies even when faced with a more complex and realistic environment, where tasks may be encountered multiple times and under varying conditions. This further highlights the adaptability and effectiveness of our method in addressing real-world challenges.

%------------------------------------------------------------------------
\section{Conclusions}
\label{sec:conclusions}
%------------------------------------------------------------------------

\noindent In this paper, we proposed a novel pipeline called TADIL for detecting and identifying tasks in task-agnostic domain-incremental learning scenarios without supervision. Our pipeline first obtains
base embeddings from the raw data using an already existing transformer-based model. The embedding densities are grouped
based on their similarity to obtain the nearest points to each cluster centroid and a task classifier is incrementally trained using only these few points. This task classifier and a drift detector are used together to learn new tasks. 

Our experiments using the SODA10M real-world driving dataset have demonstrated the good performance of the drift detector and the task classifier, and how state-of-the-art CL strategies can match the ground-truth performance when using our pipeline to predict the task ID, both in experiments assuming task boundaries using a traditional approach, and also in more realistic task-agnostic scenarios that require detecting new tasks on-the-fly.

%JG: Add some future work
%JG: done
% GB: modify with custom model
% As future work, we plan to focus on the improvement of models so that they work better in task-agnostic continual learning scenarios, by developing a reinforcement learning based component to learn a policy that helps to decide which actions should be applied to the model (e.g., apply transfer learning, use knowledge distillation, do nothing, etc.). Besides, we plan to explore other continual learning scenarios and modalities.
As future work, we aim to develop a custom Experience Replay strategy that leverages the nearest centroids by using the capability of some foundation models for zero-shot predictions to obtain weak labels for those nearest centroids without supervision. This will allow us to train the CL strategy effectively with minimal human intervention, optimizing the learning process.

% \clearpage 

\section*{Acknowledgments}
\noindent We thank Lenovo for providing the technical infrastructure to run the experiments in this paper. This work was partially supported by Lenovo and Intel as part of the Lenovo AI Innovators University Research program, by the Spanish Government under contract PID2019-107255GB-C22, and by the Generalitat de Catalunya under contract 2021-SGR-00478 and under grant 2020 FI-B 00257.

%%%%%%%%% REFERENCES
% \section{References Section}
\bibliographystyle{IEEEtran}
\bibliography{egbib}

% Generated by IEEEtran.bst, version: 1.14 (2015/08/26)
\begin{thebibliography}{10}
\providecommand{\url}[1]{#1}
\csname url@samestyle\endcsname
\providecommand{\newblock}{\relax}
\providecommand{\bibinfo}[2]{#2}
\providecommand{\BIBentrySTDinterwordspacing}{\spaceskip=0pt\relax}
\providecommand{\BIBentryALTinterwordstretchfactor}{4}
\providecommand{\BIBentryALTinterwordspacing}{\spaceskip=\fontdimen2\font plus
\BIBentryALTinterwordstretchfactor\fontdimen3\font minus
  \fontdimen4\font\relax}
\providecommand{\BIBforeignlanguage}[2]{{%
\expandafter\ifx\csname l@#1\endcsname\relax
\typeout{** WARNING: IEEEtran.bst: No hyphenation pattern has been}%
\typeout{** loaded for the language `#1'. Using the pattern for}%
\typeout{** the default language instead.}%
\else
\language=\csname l@#1\endcsname
\fi
#2}}
\providecommand{\BIBdecl}{\relax}
\BIBdecl

\bibitem{Zenke2017}
F.~Zenke, B.~Poole, and S.~Ganguli, ``{Continual Learning through Synaptic
  Intelligence},'' in \emph{Proc. 34th Int. Conf. on Machine Learning
  (ICML'17)}, ser. Proceedings of Machine Learning Research, vol.~70.\hskip 1em
  plus 0.5em minus 0.4em\relax PMLR, Aug. 6--11 2017, pp. 3987--3995.

\bibitem{Lopez-Paz2017}
D.~Lopez-Paz and M.~Ranzato, ``{Gradient Episodic Memory for Continual
  Learning},'' in \emph{Advances in Neural Information Processing Systems, vol.
  30 (NIPS 2017)}.\hskip 1em plus 0.5em minus 0.4em\relax Curran Associates
  Inc., 2017, pp. 6470--6479.

\bibitem{Rebuffi2017}
S.-A. Rebuffi, A.~Kolesnikov, G.~Sperl, and C.~H. Lampert, ``{iCaRL:
  Incremental Classifier and Representation Learning},'' in \emph{Proc. 2017
  IEEE Conf. on Computer Vision and Pattern Recognition (CVPR'17)}, Jul. 21--26
  2017, pp. 5533--5542.

\bibitem{Aljundi2018}
R.~Aljundi, F.~Babiloni, M.~Elhoseiny, M.~Rohrbach, and T.~Tuytelaars,
  ``{Memory Aware Synapses: Learning What (not) to Forget},'' in \emph{Proc.
  European Conf. on Computer Vision, ECCV 2018}.\hskip 1em plus 0.5em minus
  0.4em\relax Springer International Publishing, Sep. 8--14 2018, pp. 144--161.

\bibitem{radford2021learning}
A.~Radford, J.~W. Kim, C.~Hallacy, A.~Ramesh, G.~Goh, S.~Agarwal, G.~Sastry,
  A.~Askell, P.~Mishkin, J.~Clark, G.~Krueger, and I.~Sutskever, ``{Learning
  Transferable Visual Models From Natural Language Supervision},'' in
  \emph{Proc. 38th Int. Conf. on Machine Learning (ICML'21)}, ser. Proceedings
  of Machine Learning Research, vol. 139.\hskip 1em plus 0.5em minus
  0.4em\relax PMLR, Jul. 18--24 2021, pp. 8748--8763.

\bibitem{fusion_transformer}
A.~Prakash, K.~Chitta, and A.~Geiger, ``{Multi-Modal Fusion Transformer for
  End-to-End Autonomous Driving},'' in \emph{Proc. 2021 IEEE/CVF Conf. on
  Computer Vision and Pattern Recognition (CVPR'21)}.\hskip 1em plus 0.5em
  minus 0.4em\relax IEEE Computer Society, Jun. 19--25 2021, pp. 7073--7083.

\bibitem{motion_transformer}
Z.~Huang, X.~Mo, and C.~Lv, ``{Multi-modal Motion Prediction with
  Transformer-based Neural Network for Autonomous Driving},'' in \emph{Proc.
  39th Int. Conf. on Robotics and Automation (ICRA)}, May 23--27 2022, pp.
  2605--2611.

\bibitem{Delange_2021}
M.~De~Lange, R.~Aljundi, M.~Masana, S.~Parisot, X.~Jia, A.~Leonardis,
  G.~Slabaugh, and T.~Tuytelaars, ``{A Continual Learning Survey: Defying
  Forgetting in Classification Tasks},'' \emph{IEEE Trans. Pattern Anal. Mach.
  Intell.}, vol.~44, no.~7, pp. 3366--3385, 2022.

\bibitem{Kirkpatrick_2017}
J.~Kirkpatrick, R.~Pascanu, N.~Rabinowitz, J.~Veness, G.~Desjardins, A.~A.
  Rusu, K.~Milan, J.~Quan, T.~Ramalho, A.~Grabska-Barwinska, D.~Hassabis,
  C.~Clopath, D.~Kumaran, and R.~Hadsell, ``{Overcoming Catastrophic Forgetting
  in Neural Networks},'' \emph{Proceedings of the National Academy of
  Sciences}, vol. 114, no.~13, pp. 3521--3526, Mar. 2017.

\bibitem{PARISI201954}
G.~I. Parisi, R.~Kemker, J.~L. Part, C.~Kanan, and S.~Wermter, ``{Continual
  Lifelong Learning with Neural Networks: A Review},'' \emph{Neural Networks},
  vol. 113, pp. 54--71, 2019.

\bibitem{rolnick19}
D.~Rolnick, A.~Ahuja, J.~Schwarz, T.~Lillicrap, and G.~Wayne, ``{Experience
  Replay for Continual Learning},'' in \emph{Advances in Neural Information
  Processing Systems (NeurIPS 2019)}, vol.~32.\hskip 1em plus 0.5em minus
  0.4em\relax Curran Associates, Inc., 2019.

\bibitem{meng2022discl}
M.~Mirza, M.~Masana, H.~Possegger, and H.~Bischof, ``{An Efficient
  Domain-Incremental Learning Approach to Drive in All Weather Conditions},''
  in \emph{Proc. 2022 IEEE/CVF Conf. on Computer Vision and Pattern Recognition
  Workshops (CVPRW'22)}, Jun. 19--20 2022, pp. 3000--3010.

\bibitem{cl_medical_seg}
\BIBentryALTinterwordspacing
C.~Gonz{\'{a}}lez, G.~Sakas, and A.~Mukhopadhyay, ``What is wrong with
  continual learning in medical image segmentation?'' \emph{CoRR}, vol.
  abs/2010.11008, 2020. [Online]. Available:
  \url{https://arxiv.org/abs/2010.11008}
\BIBentrySTDinterwordspacing

\bibitem{xie2022general}
J.~Xie, S.~Yan, and X.~He, ``{General Incremental Learning with Domain-aware
  Categorical Representations},'' in \emph{Proc. 2022 IEEE/CVF Conf. on
  Computer Vision and Pattern Recognition (CVPR'22)}.\hskip 1em plus 0.5em
  minus 0.4em\relax IEEE Computer Society, Jun. 21--24 2022, pp.
  14\,331--14\,340.

\bibitem{shin2022task}
H.~Zhu, M.~Majzoubi, A.~Jain, and A.~Choromanska, ``{TAME: Task Agnostic
  Continual Learning using Multiple Experts},'' 2022, arXiv preprint
  arXiv:2210.03869.

\bibitem{Shin_2017}
H.~Shin, J.~K. Lee, J.~Kim, and J.~Kim, ``{Continual Learning with Deep
  Generative Replay},'' in \emph{Advances in Neural Information Processing
  Systems, vol. 30 (NIPS 2017)}.\hskip 1em plus 0.5em minus 0.4em\relax Curran
  Associates Inc., 2017, pp. 2994--3003.

\bibitem{li2018learning}
Z.~Li and D.~Hoiem, ``{Learning without Forgetting},'' \emph{IEEE Trans.
  Pattern Anal. Mach. Intell.}, vol.~40, no.~12, pp. 2935--2947, 2018.

\bibitem{schwarz2018progress}
J.~Schwarz, W.~Czarnecki, J.~Luketina, A.~Grabska{-}Barwinska, Y.~W. Teh,
  R.~Pascanu, and R.~Hadsell, ``{Progress {\&} Compress: {A} Scalable Framework
  for Continual Learning},'' in \emph{Proc. 35th Int. Conf. on Machine
  Learning, (ICML'18)}, ser. Proceedings of Machine Learning Research,
  vol.~80.\hskip 1em plus 0.5em minus 0.4em\relax PMLR, Jul. 10--15 2018, pp.
  4535--4544.

\bibitem{Aljundi_2019}
R.~Aljundi, M.~Lin, B.~Goujaud, and Y.~Bengio, ``{Gradient Based Sample
  Selection for Online Continual Learning},'' in \emph{Advances in Neural
  Information Processing Systems, vol. 32 (NeurIPS 2019)}, Dec. 8--14 2019, pp.
  11\,816--11\,825.

\bibitem{rusu2016progressive}
A.~A. Rusu, N.~C. Rabinowitz, G.~Desjardins, H.~Soyer, J.~Kirkpatrick,
  K.~Kavukcuoglu, R.~Pascanu, and R.~Hadsell, ``{Progressive Neural
  Networks},'' 2022, arXiv preprint arXiv:1606.04671.

\bibitem{li2017learning}
Y.~Li, J.~Yang, Y.~Song, L.~Cao, J.~Luo, and L.~Li, ``{Learning from Noisy
  Labels with Distillation},'' in \emph{Proc. 2017 IEEE Int. Conf. on Computer
  Vision (ICCV)}.\hskip 1em plus 0.5em minus 0.4em\relax IEEE Computer Society,
  Oct. 22--29 2017, pp. 1928--1936.

\bibitem{mallya2018packnet}
A.~Mallya and S.~Lazebnik, ``{PackNet: Adding Multiple Tasks to a Single
  Network by Iterative Pruning},'' in \emph{Proc. 2018 IEEE/CVF Conf. on
  Computer Vision and Pattern Recognition (CVPR'18)}.\hskip 1em plus 0.5em
  minus 0.4em\relax IEEE Computer Society, Jun. 18--23 2018, pp. 7765--7773.

\bibitem{he2016deep}
K.~He, X.~Zhang, S.~Ren, and J.~Sun, ``{Deep Residual Learning for Image
  Recognition},'' in \emph{Proc. 2016 IEEE Conf. on Computer Vision and Pattern
  Recognition (CVPR'16)}, Jun. 27--30 2016, pp. 770--778.

\bibitem{tibshirani2002diagnosis}
R.~Tibshirani, T.~Hastie, B.~Narasimhan, and G.~Chu, ``{Diagnosis of Multiple
  Cancer Types by Shrunken Centroids of Gene Expression},'' \emph{Proceedings
  of the National Academy of Sciences of the United States of America},
  vol.~99, no.~10, pp. 6567--6572, 2002.

\bibitem{han2021soda10m}
J.~Han, X.~Liang, H.~Xu, K.~Chen, L.~Hong, J.~Mao, C.~Ye, W.~Zhang, Z.~Li,
  X.~Liang, and C.~Xu, ``{SODA10M: A Large-Scale 2D Self/Semi-Supervised Object
  Detection Dataset for Autonomous Driving},'' 2021, arXiv preprint
  arXiv:2106.11118.

\bibitem{verwimp2022clad}
E.~Verwimp, K.~Yang, S.~Parisot, L.~Hong, S.~McDonagh, E.~Pérez-Pellitero,
  M.~{De Lange}, and T.~Tuytelaars, ``{CLAD: A Realistic Continual Learning
  Benchmark for Autonomous Driving},'' \emph{Neural Networks}, vol. 161, pp.
  659--669, 2023.

\end{thebibliography}

\vfill

\end{document}